\documentclass[journal, twoside]{ieeetran}
\pdfoutput=1 
\IEEEoverridecommandlockouts                              %

\usepackage{graphics} %
\usepackage{epsfig} %
\usepackage{mathptmx} %
\usepackage{times} %
\usepackage{amsmath} %
\usepackage{amssymb}  %
\usepackage{url}
\usepackage{diagbox} 
\usepackage{float}%
\usepackage{xcolor, soul}
\usepackage{graphicx}
\usepackage{bm}
\usepackage{float}
\usepackage{multirow}
\usepackage{makecell}
\usepackage[ruled,linesnumbered]{algorithm2e}
\usepackage{scalerel}
\usepackage{tikz}
\usepackage[labelfont=bf]{caption}
\usepackage{subcaption}
\usepackage{pifont}
\usepackage{booktabs}
\usepackage{multirow}

\newcommand{\revcolor}[1]{}

\usepackage[
    natbib=true,
    style=ieee,
    sorting=none
]{biblatex}

\usetikzlibrary{svg.path}

\definecolor{orcidlogocol}{HTML}{A6CE39}
\tikzset{
  orcidlogo/.pic={
    \fill[orcidlogocol] svg{M256,128c0,70.7-57.3,128-128,128C57.3,256,0,198.7,0,128C0,57.3,57.3,0,128,0C198.7,0,256,57.3,256,128z};
    \fill[white] svg{M86.3,186.2H70.9V79.1h15.4v48.4V186.2z}
                 svg{M108.9,79.1h41.6c39.6,0,57,28.3,57,53.6c0,27.5-21.5,53.6-56.8,53.6h-41.8V79.1z M124.3,172.4h24.5c34.9,0,42.9-26.5,42.9-39.7c0-21.5-13.7-39.7-43.7-39.7h-23.7V172.4z}
                 svg{M88.7,56.8c0,5.5-4.5,10.1-10.1,10.1c-5.6,0-10.1-4.6-10.1-10.1c0-5.6,4.5-10.1,10.1-10.1C84.2,46.7,88.7,51.3,88.7,56.8z};
  }
}
\newcommand\orcidicon[1]{\href{https://orcid.org/#1}{\mbox{\scalerel*{
\begin{tikzpicture}[yscale=-1,transform shape]
\pic{orcidlogo};
\end{tikzpicture}
}{|}}}}

\makeatletter
\let\NAT@parse\undefined
\makeatother
\usepackage[colorlinks=true,linkcolor=blue,citecolor=blue]{hyperref} %

\hypersetup{
  colorlinks   = true,    %
  urlcolor     = blue,    %
  linkcolor    = blue,    %
  citecolor    = blue      %
}

\title{\LARGE \bf
Active Multi-Object Exploration and Recognition \\via Tactile Whiskers
}
\author{Chenxi Xiao \orcidicon{0000-0002-7819-9633}, Shujia Xu \orcidicon{0000-0002-3173-8249}, Wenzhuo Wu \orcidicon{0000-0003-0362-6650}, and Juan Wachs$^*$ \orcidicon{0000-0002-6425-5745}%
\thanks{ Chenxi Xiao is with the School of Industrial Engineering  at Purdue University, 
        {\tt\small xiao237@purdue.edu}}%
\thanks{ Shujia Xu is with the School of Industrial Engineering at Purdue University, 
        {\tt\small xu1377@purdue.edu}}%
\thanks{ Wenzhuo Wu is with the School of Industrial Engineering at Purdue University, 
        {\tt\small wenzhuowu@purdue.edu}}%
\thanks{ Juan Wachs (*corresponding author) is with the School of Industrial Engineering   at Purdue University, 
        {\tt\small jpwachs@purdue.edu}}%
}

\begin{document}

\maketitle
\thispagestyle{empty}
\pagestyle{empty}

\maketitle

\begin{abstract}
Robotic exploration under uncertain environments is challenging when optical information is not available. In this paper, we propose an autonomous solution of exploring an unknown task space based on tactile sensing alone. We first designed a whisker sensor based on MEMS barometer devices. This sensor can acquire contact information by interacting with the environment non-intrusively.

This sensor is accompanied by a planning technique to generate exploration trajectories by using mere tactile perception. This technique relies on a hybrid policy for tactile exploration, which includes a proactive informative path planner for object searching, and a reactive Hopf oscillator for contour tracing. Results indicate that the hybrid exploration policy can increase the efficiency of object discovery. 

Last, scene understanding was facilitated by segmenting objects and classification. A classifier was developed to recognize the object categories based on the geometric features collected by the whisker sensor. Such an approach demonstrates the whisker sensor, together with the tactile intelligence, can provide sufficiently discriminative features to distinguish objects.
\end{abstract}


\section{Introduction}

\IEEEPARstart{T}{actile} sensing can augment, complement, and sometimes substitute vision when dealing with occluded and concealed objects \cite{nichols2015methods}, transparent or highly reflective materials \cite{taira20103d}, and when handling liquids \cite{nagai2020tactile}, for which optical sensing is not well suited. An example application scenario is to explore objects buried or underwater, in which operators rely on tactile perception to localize and then acquire object properties. {\revcolor{red} When optical information is limited or not available, we propose an autonomous system to localize, characterize and recognize objects based on tactile sensing alone. The goal is to gather information efficiently while not perturbing the surroundings. Inspired by human’s tactile exploratory behavior, here we propose strategies and methods for autonomous exploration in a cluttered scene. The proposed technologies facilitate reconstructing an occupancy map, while simultaneously localizing and recognizing the objects in a human-like fashion.  In previous works, most exploratory tasks in robotics 
were conceived through remote sensing (e.g., vision \cite{schmid2020efficient}, sonar \cite{yordanova2020coverage}, etc.). That line of work proposed the collection of observations from a contact-free space, where the contact events were commonly avoided altogether. This is mainly because contact events can be intrusive, leading objects to be repositioned, and in turn, increasing the uncertainty of historic observations. Another challenge is the discontinuities in control and decision-making resulting from contact events. More specifically, while a variety of tactile exploration strategies have been studied \cite{driess2017active, rosales2018gpatlasrrt, matsubara2016active}, most solutions for surface exploration assume that there is a single object represented in terms of a watertight surface. The strategy for generalizing to explorations in cluttered scenes, and for non-continuous surfaces has been rarely studied.%

\begin{figure}[t]
    \centering
    \includegraphics[width=\linewidth]{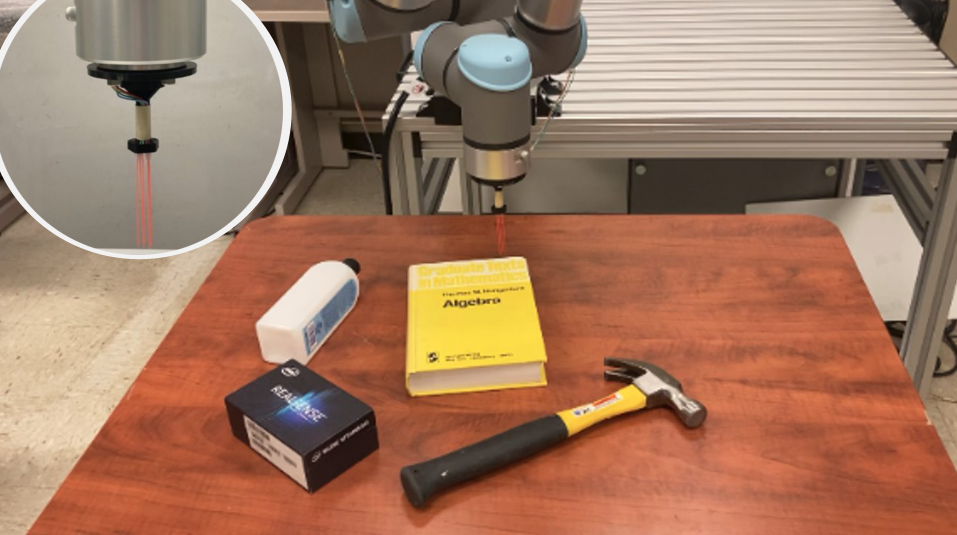}
    \caption{A robot is using the tactile feedback from our developed whisker sensor to localize, and recognize the objects found during exploration. The video is available at: {\url{https://youtu.be/qB5x9PDLV0o}}}%
    \label{fig:teaser}
\end{figure}
We propose a hybrid exploration policy, in which a reactive contour tracing policy is integrated for the first time. In this design, objects are localized by an informative path planner, which is used to plan an exploration path to reduce the uncertainty within a scene. When a new object is discovered, its contour is then traced in order to collect contact points. This process is reactive, which minimizes the disruption to the environment. Our work also shows that contour tracing is a greedy policy which boosts the exploration efficiency measured over the traveling distance, and provides this methodology to integrate contour tracing policy into a general informative path planning framework. By comparing to two baselines, we achieved a larger amount of information gathered, better point coverage on the shape, and a significant improvement in the average travel distance.

Further, the intrusiveness during interaction is reduced by introducing a whisker array of tactile sensor in the form of filaments. Whiskers are flexible, allowing compliant interaction with minimal contact forces \cite{pearson2011biomimetic, fox2012tactile, lepora2019pixels}. In comparison to conventional tactile sensors, whisker filaments allow larger exploration areas due to their length, and form factor. This enables object detection without the need to make full surface contact. The receptor at the distal end of whiskers is highly sensitive, allowing the detection of micro-force variations such as mechanical vibration and airflow \cite{arabzadeh2004whisker, yan2019whisker}. These features make whiskers particularly suitable for spatial exploration tasks. In this paper, a barometer sensor based whisker sensor was developed, and demonstrated on active tactile exploration. This sensor consists of: (1) plastic filaments used for detecting contacts, and (2) receptor units that measure the pressure caused by filament bending. Compared to the previous work, our hardware has a variety of improvements, including MEMS barometers of higher sensitivity, higher maximum pressure range, compact size, and scalability to larger arrays. The common issue of barometer's drifting issue is addressed, and we leverage an active sweeping motion to augment contact signals.

Last, a new classification paradigm is proposed based on the estimated object contour to make sense of the observations collected by whiskers. In contrast to most common object classifiers which require either high-resolution visual images or point clouds with hundreds of points, our approach only requires a minimum number of points (e.g., 8-30 points per object). Using contour shapes allows satisfactory classification results (e.g., 98.3\% accuracy for 11 real objects). The proposed combination of sensing and exploration reduces the time and energy cost required in the state-of-the-art tactile sampling. In addition, the proposed classifier is generalizable to 3D point clouds when incorporating surface contacts samples.  

Our technical contributions are listed as follows.  
\begin{enumerate}
    \item A low-cost whisker array tactile sensor designed for non-intrusive active tactile exploration (Sec.~\ref{sec:wsd}).
    \item A hybrid autonomous tactile exploration policy for both searching objects and tracing contours (Sec.~\ref{sec:te}).  
    \item Object categories classification using very few contact points (Sec.~\ref{sec:recog})
    \item Evaluation experiments on whiskers, exploration policy, and classification (Sec.~\ref{sec:res_whisker}, Sec.~\ref{sec:res_te}, Sec.~\ref{sec:res_or}).
\end{enumerate}

}

\section{Related Work}

\subsection{Tactile Sensors}
Tactile sensors are devices that acquire tactile information through physical interaction with the environment. The most common tactile sensors are based on capacitive, piezoresistive, thermoresistive, inductive, piezoelectric, magnetic, and optical sensing mechanisms \cite{tiwana2012review}. {\revcolor{red} While each sensing mechanism only provides a specific object attribute, recent progress in the design and manufacturing of novel tactile sensors can capture broader types of attributes as opposed to their predecessors. For instance, BioTac, a commercially available tactile sensor can convey contact features that include pressure, vibration, temperature, and material conductivity \cite{fishel2012sensing}. Likewise, tactile sensors based on local optical imaging (e.g., Gelsight \cite{yuan2017gelsight}, Soft-Bubble \cite{alspach2019soft}, Digit \cite{lambeta2020digit}, and TacTip \cite{ward2018tactip}) convey optical attributes. Additional characteristics such as 3D geometric can be predicted using photometric stereo reconstruction and machine learning methods, which express well the correlation of the available information and what is known from previous experience \cite{yuan2017gelsight}. With the help of these techniques and others, tactile sensing is becoming more comparable between humans and robotics.}

Our focus is on tactile sensors that are suitable for spatial exploration tasks. However, most conventional tactile sensors only have a very narrow sensing range and thus lack efficiency. For example, the commercial force-sensitive resistors (FSR), piezoresistive, and capacitive sensors are all manufactured into thin films (or plates), which constrains the contact to be inside a local surface region \cite{schofield2016effect, maiolino2013flexible}. To address this limitation, we developed a whisker tactile sensor, which has a wide sensing range and high sensitivity. We report on satisfactory results in exploring a variety of challenging scenes.

\subsection{Whisker Sensors}

Whiskers (Vibrissae) sensing is commonly found in aquatic mammals, rodents, insects \cite{prescott2011vibrissal, ahl1986role}, and even viruses \cite{kostyuchenko2005tail}, in which hair-like or bristle-like structures are used for the tactile perception of the surroundings. Whisker sensing endows nature with a variety of functions. The whiskers allow enclosure, compliance, separation, heat dissipation, navigation, and wave propagation  \cite{pearson2011biomimetic}. 

Artificial whiskers, which are inspired by the biological structure of the vibrissae, have been adopted to enhance the sensing capabilities of robots. The earliest whisker sensor can be traced back to the 1970s \cite{wang1978sensors}, and more recently they have been used for applications including obstacle avoidance \cite{mckerrow1991introduction}, ground proximity \cite{hirose1985titan}, and object localization \cite{fox2012tactile}, etc.

What makes whiskers so effective in the animal kingdom is the quality and quantity of the sensory information obtained. These advantages can be attributed to the whisker's sensing range, high sensitivity, and compliance. For example, the whiskers on a rat’s face have a density of around 30 on each side of the face, and lengths ranging from 20-100 mm \cite{brecht1997functional}, which enables detecting small objects in a wide range around the rat's head. 
Similarly, these properties are also preferred on robots, due to the need of exploring the distribution of objects ahead of time. To reach information acquisition efficiency that is similar to animals, efforts have been devoted to fabricating whiskers as an array \cite{harada2014fully, fend2003active}. More recently, Struckmeier et al. \cite{struckmeier2019vita} improved the sensing mechanism by reproducing the active whisking behavior that is observed in rodents by motor ``musculature'', which is capable of delivering a sweeping motion. {\revcolor{red} TacWhisker \cite{lepora2018tacwhiskers} leverages optical imaging to visualize the locations of whisker's distal points. This allows to scale up the whisker density with only an individuate receptor unit.} The work most relevant to our sensor is the lightweight whisker array designed by Deer et al. \cite{deer2019lightweight}, which shows barometers with extended whiskers can be used as the sensing component to detect micro-force  (e.g. detect air or fluid velocity around whiskers). Based on this sensing mechanism, various adaptions in mechanical, electrical, and signal processing aspects have been made to facilitate the usage of barometer based whiskers in active tactile exploration (refer to Sec.~\ref{sec:wsd}).

\subsection{Active Tactile Spatial Exploration} \label{sec:rw:tse}
Active spatial exploration concerns the acquisition of the scene or object's spatial features using active exploration. The spatial exploration allows the agent to gather the information necessary to address challenges such as scene reconstruction \cite{driess2017active}, object recognition \cite{zhang2017active}, pose estimation \cite{suresh2020tactile}, and planning manipulation policies \cite{dragiev2013uncertainty}, which are stepstones in realistic applications leveraging robotics and machine perception.

Most active spatial exploration problems rely on the visual sensing modality. For example, the Active Simultaneous Localization and Mapping (Active SLAM) uses an active policy to guide the map reconstruction while simultaneously localizing the robot's pose \cite{carrillo2012comparison}. The same problem is more challenging when conditioned solely on the tactile modality, which is yet less investigated. When compared with the visual modality,  the tactile sensing range is much shorter, resulting in reduced efficiency in information acquisition. For the same reason, each contact is generally not sufficiently informative about the object's properties.  While this problem can be alleviated by accumulating evidence from a large number of probes, there is a  movement cost associated with the finger transitions \cite{zhang2017active}. 
Besides, in some applications like bomb disposal, every probing motion could also lead to fatal outcomes. 

To sample using an optimal strategy with a reduced number of probes, motion planners for information acquisition have been previously proposed. This is part of a research theme referred to as Informative Path Planning (IPP)  \cite{hitz2017adaptive, schmid2020efficient, yordanova2020coverage}. Among IPP approaches developed for tactile sensors, a commonly used approach is to explore uncertain regions by adding discrete samples. For example, the next probing point can be determined by solving the Bayesian optimization problem on a continuous function \cite{srinivas2009gaussian}.  Other approaches have also focused on developing efficient sampling policies to accelerate uncertainty reduction, and has shown to be able to increase tactile sampling efficiency by Jamali et al. (2016) \cite{jamali2016active}, Martinez et al. (2017) \cite{martinez2017feeling}, and Kaboli et al. (2019) \cite{kaboli2019tactile}. The downside of such approaches is that discrete probes convey no observations during the transition between two probing events. 

Conversely, sliding or sweeping motion (i.e., sample continuously along the object surface) can generate more efficient exploration paths \cite{driess2017active}. For example, it has been shown that continuous informative tactile sampling can be achieved through a sliding motion on flat surfaces \cite{abraham2017ergodic}. More recently, sliding motion has also been demonstrated on curved object surfaces. For instance, Driess et al. \cite{driess2017active, driess2019active} used a compliant controller to facilitate data collection along object surfaces, and simultaneously used Gaussian Process Regression to estimate the object shape. Similar methods have also been adopted by Rosales et al. (2018) \cite{rosales2018gpatlasrrt} and Ottenhaus et al. (2018) \cite{ottenhaus2018active}.  While the works mentioned above indicate that simple geometric shape estimation can be accomplished through sliding sensor motion, the effectiveness of such an approach on complex object surfaces remains to be shown. {\revcolor{red} The challenges lie in both planning and control. The planned trajectory has only a limited horizon due to the high uncertainty on incomplete surfaces. In addition, an abrupt change on the surface curvature (e.g., stairs) could induce intrusive forces if the contact normal force is not measured accurately and timely (e.g., due to multiple contact points or friction force).}

\subsection{Tactile Object Recognition}

Tactile object recognition is an essential component to achieve tactile intelligence, which is the ability of machines to make sense of the observations based on tactile sensing \cite{luo2017robotic}. In most cases, the sensor configuration would determine the modality attributes used in the recognition task. Therefore, the specific design will be more sensitive to one or multiple options from geometry, texture, material stiffness, mass, etc. than others \cite{tiwana2012review}. Particularly, geometric shape is a commonly used feature since it is one of the most intuitive object representations and can be easily measured with tactile sensing. {\revcolor{red} In this category, recognition is based on local shape, such as  shape descriptors including LBP \cite{li2013sensing}, SIFT \cite{lowe1999object} and Tactile-SIFT \cite{luo2015novel}, MR-8 \cite{varma2005statistical}, Normalized momentum \cite{pezzementi2011tactile}, etc. Using an optical tactile sensor allows obtaining high resolution local tactile images. This facilitates object discrimination with similar dimensions (such as coins) \cite{abad2020low}. Another branch of work is using global features to characterize objects. For example, point cloud models can be obtained by sampling the whole object surface. A point cloud model can be converted to a high resolution watertight surface (namely implicit surface) \cite{driess2017active}. This enables object visualization and occupancy information \cite{gandler2020object}. In addition, volume, and the global point distribution can be matched to help discriminate objects \cite{meier2011probabilistic, casselli1995robustness}.

In this paper, we propose a novel approach that can be used to recognize objects by the projected contour. Compared to other shape attributes, the contour is a “cheap” proxy based on the polygon representation. This is particularly suitable for representing tactile observations. It can be easily obtained by contour tracing, and requires fewer computational resources than volumetric attributes. We conducted a number of experiments to show that the contour shape is sufficient to allow object classification over a number of categories.}

\section{Whisker Sensor Design} \label{sec:wsd}

A prototype of a novel artificial whisker sensor is presented. This device consists of three parts: 1) whisker filament, which was created by plastic-based soft materials that can propagate the contact force, and 2) a barometer based pressure sensing device which works as the ``receptor'', and 3) a programmable microprocessor for data processing and communication.  {\revcolor{red}Note that, using a barometer sensor as the signal receptor enables high pressure sensitivity, which in turn, allows the detection of minor contact forces without disruption to the environment.}

\begin{figure}
    \centering
    \includegraphics[width=\linewidth]{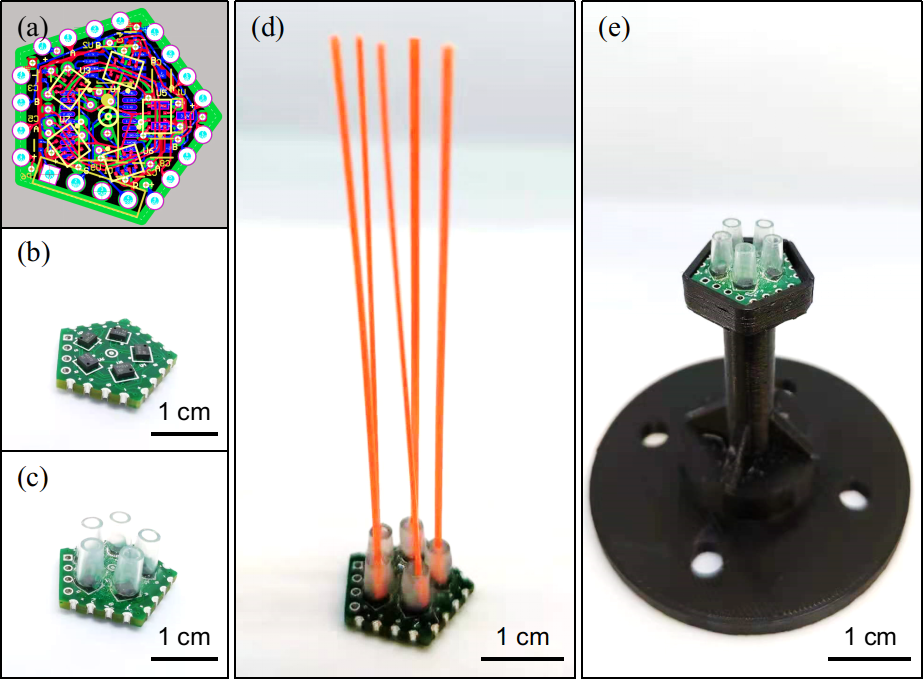}
    \caption{The fabrication procedure of the whisker sensor. (a) PCB layout, (b) sensor board, (c) install plastic tubes, (d) insert whiskers, (e) fix to the tool flange.}
    \label{fig:hardware_system}
\end{figure}

Five barometers were soldered on the top surface of a Printed Circuit Board (PCB), as shown in Fig.~\ref{fig:hardware_system}. {\revcolor{red}Each PCB board was designed in a pentagon shape to allow interfacing with other boards seamlessly. This feature can be utilized to build a larger whisker array, which can be arranged as a partial surface with an icosahedron shape (e.g., soccer ball).} Plastic tubes with a diameter of 6 mm and a height of 7 mm were then fixed on the barometers using epoxy resin (Devcon). {\revcolor{red} The size of the tube fits the barometer unit's size, which ensures the barometer to be in an airtight chamber.} Ecoflex 00-30 (Smooth-On) was applied to connect the whiskers with the barometers. Part A and part B of Ecoflex were mixed uniformly at a weight ratio of 1:1 and degassed for 10 min under a vacuum condition. Then, the Ecoflex was injected into the plastic tubes using a syringe. Each whisker with a diameter of 0.8 mm and a length of 7 cm was then inserted into a plastic tube. {\revcolor{red}This diameter provides sufficient stiffness to propagate the deformation of the whisker tip to the gel layer, and allows good sensitivity to detect contact events. While a wide range of diameters could be applied as well, smaller diameters (less than 0.25mm) reduce the signal magnitude significantly, making contact detection a more challenging task. In terms of lengths, longer lengths can facilitate a larger detection range. However, this would lead to a larger positional offset when in contact, which may decrease the accuracy of contact localization.} Next, the whiskers were fixed horizontally with the assistance of a supporting skeleton. Finally, the PCB with fixed whiskers were cured in an oven at 65 $^{\circ}$C for an hour to crosslink the Ecoflex. After curing, the whole PCB was attached to the tool flange (3D printed using PLA).

One DPS310 digital barometer (Infineon Technologies) was used to sense the pressure at each whisker's root. {\revcolor{red}This barometer provides 24-bits digital measurement readings. As a result, the sensitivity is sufficiently high to detect contact events on the whisker tip, even without removing the air within barometer's case. The sensitivity can be further improved by injecting the gel into the case.} The sampling rate of all barometers was configured to 64 Hz, {\revcolor{red}with oversampling rate as 8. This is a tradeoff between speed and signal-noise ratio}. To control and read data from 5 barometers, a low-cost, 8-bits microprocessor STM8S003F3P6 (ST Microelectronics) was used. The SPI bus enabled serial communication between the microprocessor and the onboard sensors. 

The system also consists of a computer that {\revcolor{red}queries the sensor readings, which is the same computer that runs the algorithm described in the following sections}. RS-485 serial communication was used for
real-time communication between the sensor board and the computer. The RS-485 only has two wires, and allows up to 256 sensor boards on a single bus segment. Using communication, the sensor board is scalable to a large sensor array, as each pentagon board has 4 pairs of communication ports, which can be connected to any other sensor board seamlessly.  
To eliminate the data packet collision in transmission, a communication protocol was designed based on token-ring. For this, the transmission of sensor $S_0$ is triggered by the computer. The transmission of sensor $S_k$ ($k \geq 1$) can only be triggered by sensor $S_{k-1}$.

Pressure drifting is a commonly observed issue in barometers. This is mainly caused by the coupling effects from the temperature, and the partial inelastic strain recovery of the gel when a large force is applied \cite{koiva2020barometer, deer2019lightweight}. 
The drifting is harmful because it reduces the signal-noise ratio, and thus may degrade the confidence in detecting contacts.  
The drifting can be removed by a high pass filter because it mainly contains low frequent components. For this, a first-order high pass filter with the cutoff frequency at 11.3 Hz was used to remove the drifting effect. The output signal of this high pass filter was then rectified to be positive. The noise was then removed by a first-order low pass filter with the cutoff frequency at 31.8 Hz. {\revcolor{red}After the filter, no significant variation is observed in each channel's output. By keeping the sensor in a stationary environment for approximately three and a half hours, the standard deviation of the sensor's output (averaged over all channels) is measured to be $2.2\times10^{-4}$ kPa, with mean value at $9.5\times10^{-7}$ kPa}. A contact event could be recognized if the filter output exceeds a threshold (0.001 kPa in our case).

A special case is that the high-frequency signal component does not exist when the whisker rod is attached to the object in a stationary state. We solve this by adding a rotatory motion to the robot end-effector's z-axis. This strategy is known as ``whisking'' in animals, which provides the necessary excitatory input to identify static objects \cite{sofroniew2015whisking}. In our case, a sinusoid motion of $\pm0.5$ degree magnitude at 0.5 Hz was applied. {\revcolor{red}This motion magnitude can effectively augment a contact signal while not causing variations to the sensor's reading when in the contact-free state.}

\section{Autonomous Tactile Exploration} \label{sec:te}

\subsection{Problem Formulation}

Consider a dexterous end-effector with tactile sensing ability. A contact point  $\mathbf{p}_t$ in the world coordinate can be obtained when an object is in contact with the tactile sensor. As an example, refer to the scene in Fig.~\ref{fig:teaser}, in which the task space is defined inside a rectangular region in a horizontal plane: $z=h_d$, {\revcolor{red}where $h_d$ is the height value for this plane specified by the user},  $x \in [x^-, x^+]$, $y \in [y^-, y^+]$. The boundaries of this region can be relaxed to an irregular polygon. Also, the $z=h_d$ plane can be relaxed to a smoothed surface $z=f(x,y)$ if the end-effector can slide to explore such surface. It is also a common case that only the end-effector is equipped with the tactile sensor. To avoid the case that the object may be unintentionally moved by the robot regions that are not equipped with tactile skin, only the tactile sensor is placed inside the task space.

Inside the robot's task space $\mathbb{T}$, there exists an unknown number of objects $\mathbb{O}_k, k \in 1, 2, ..., N_o$, each with its contour curve $C_k, k \in 1, 2, ..., N_o$. For any point pairs $\bm{p}_{C_{i}} \in C_i, \bm{p}_{C_{j}} \in C_j, i \neq j$ , it satisfies the spatial isolation condition:
\begin{equation}
    ||\bm{p}_{C_{i}} - \bm{p}_{C_{j}}|| > D_s 
    \label{eq:isolation}
\end{equation}

Where $D_s$ is a distance constraint that assures objects do not attach to each other, and that sensor is able to move and explore freely between objects. This is a sufficient condition that allows the objects to be spatially separated by using only tactile observations.

The goal of tactile exploration is to localize and spatially characterize each object. While this can be achieved through exhaustive exploration, it would not be practically efficient motion-wise. The reason is that every single ``touch'' involves costs expressed in terms of travel distance, movement time, energy, and computing cost.  Compared to visual exploration,  we generally attempt to be efficient in tactile exploration, due to the costs and the limited amount of information conveyed per ``touch''. Without loss of generality, this also reduces the chance of target repositioning by reducing the number of ``touches'' if intrusive tactile sensors are used.

For this reason, the aim is to minimize the traveling cost $\int_{p} c(s) ds$ along the exploration path $p$ starting at the robot's current position. Hereby, we define the acquisition function at time step $t$ as $h^t(\textbf{x})$. Assume the traveling time of path $p$ is $T$, the corresponding optimization problem is defined as Eq.~(\ref{eq:optim1}):

\begin{equation}
\begin{array}{ll}
\underset{\pi}{\operatorname{minimize}} & \int_{p} c(s) ds  \\
\text { subject to } & h^{T}(\bm{p}_{C_i})<H \ \ \forall \bm{p}_{C_i} \in C_i \\
& i = 1, 2, ..., N_o
\end{array}
\label{eq:optim1}
\end{equation}

Where $H$ is a given threshold value to ensure the object boundaries are sufficiently explored. This is subject to the Informative Path Planning (IPP) framework, on the region around the object contours rather than the whole task space. Obtaining analytical solutions for this problem is NP-hard (Feige et al. \cite{feige1998threshold}). {\revcolor{red}Also, the constraint condition is defined on $C_i$, which can only be calculated when the groundtruth contour is known in advance}. To make the computation tractable, another problem description is generally being solved to obtain suboptimal solutions i.e., maximizing the information acquisition within cost budgets.

\begin{equation}
\begin{array}{ll}
\underset{\pi}{\operatorname{maximize}} & \int_{p} h^0(s) ds, \\
& s \in C_i,\ i = 1, 2, ..., N_o \\
\text { subject to } & \int_{p} c(s) ds = c_t
\label{eq:problem}
\end{array}
\end{equation}

The problem of planning path $p$ is characterized by the unique properties of tactile sensing. There are two key differences with the traditional optical sensing based IPP frameworks: 1) the sensing range is the same as the contact range, leading to unavoidable contacts with objects; 2) observations are associated with motion constraints, which are incrementally added to the task space as the contact occurs. Although versatile informative path planners have been proposed to generate continuous exploration paths, most of them are either designed for contact-free task space \cite{binney2010informative, hitz2017adaptive}, or for obstacles detected ahead of time \cite{schmid2020efficient, wei2020informative}. Such algorithms may not be capable to tackle scenarios with unpredictable contacts. To collect samples from the object surface, one adaption is to follow a re-planned path using partial observations, which is re-instantiated every time a new contact occurs. But there are additional issues related to the motion consistency and exploration efficiency. Besides, making contacts in this pattern is computationally inefficient due to frequent interruptions, which involves more computation cycles required by path re-planning. In addition, these approaches may not have theoretical guarantees for full coverage of the exploration region (i.e., cannot ensure object discovery). To address all these issues, a hybrid exploration policy is introduced in the next section.

\subsection{Hybrid Exploration Policy} \label{sec:hybrid}

A hybrid tactile exploration policy is proposed to tackle the aforementioned shortcomings of the presented techniques. The rationale behind this idea is adapted from the blind's tactile exploration strategies (e.g., search objects blindly by haptics, or explore a tactile symbol image for comprehension), which relies on multiple observations or simultaneous force feedback. According to the human study from Zhang et al. (2018) \cite{zhang2018image}, this exploration procedure is characterized by five different procedures, each exhibiting a distinct motion pattern. These motion patterns are: 1) Frame Following (FF), which traces the scene boundary to obtain its size, 2) Contour Following (CF), to learn object's size and shape, 3) Surface Swiping (SS), which explores object's internal structure, 4) Relative (RE) and 5) Absolute (AB), which obtain the object's relative and absolute position by moving a finger back-and-forth, respectively. 

\begin{figure}
    \centering
    \subfloat[]{\includegraphics[width=0.3\linewidth]{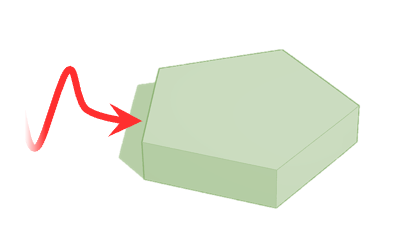}}
    \subfloat[]{\includegraphics[width=0.3\linewidth]{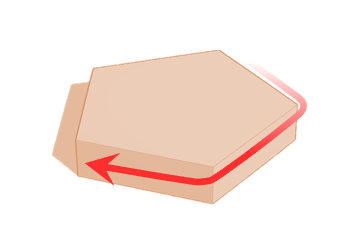}}
    \subfloat[]{\includegraphics[width=0.3\linewidth]{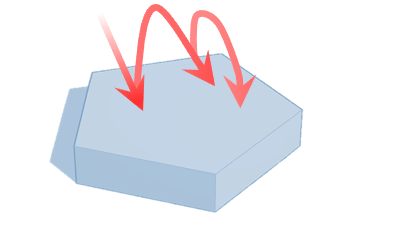}}
    \caption{Example exploration trajectory of (a) Object Searching (OS), (b) Contour Tracking (CT), and (c) Feature Sampling (FS).}
    \label{fig:procedure_examples}
\end{figure}

Let us define tactile exploration procedures of a robotic system by three independent procedures: 1) Object Searching (OS), which is used to actively search and localize an object, 2) Contour Tracing (CT), which explores the object contour, and 3) Feature Sampling (FS) that is used for actively gathering object features. Examples of such procedures are given in Fig.~\ref{fig:procedure_examples}. The OS and CT are adapted from the human blind's exploration strategies mentioned above. For these two stages, the corresponding policies are discussed in Sec.~\ref{sec:ct} and Sec.~\ref{met:os}. In addition, there is an FS procedure to collect additional information required by the task  (Sec.~\ref{sec:res_or_problem_def}). As opposed to human blind's exploration, procedures RE, AB are not always involved due to the availability of accurate positioning of robotics, leading to a reduced number of steps when compared to the human counterpart. Conversely, human blind exploration relies on kinesthetic and cognitive motor functions to obtain the location of objects \cite{zhang2018image, weaver2007attention}, which consists of the RE and AB procedures {\revcolor{red} to construct a cognitive model of a scene by using the absolute and relative location of objects}.

\subsection{Contour Tracing} \label{sec:ct}

The contour tracing is the most commonly found motion for object recognition performed by blind people \cite{zhang2018image}. {\revcolor{red}The key insight presented in this section is that contour tracing allows to boost the exploration efficiency greedily by having additional information about the occupancy state being enclosed by the path, which is proved as follows. }

\textbf{Proposition 1}: Contour tracing brings a higher information acquisition rate than a \textit{pure exploration policy} (an iterative process that collects information about areas that have not been explored up to that iteration \cite{shyam2019model}) for closed tracing paths.

\textbf{Proof}: Consider a robot exploring the task space $\mathbb{T}$ by tracing the contour of an object using path $p$. Since the object has not been visited before, it is labeled as ``not occupied'' in an equivalent grid representation. A pure exploration policy has an information acquisition rate $\frac{\int_p h(s) ds}{\int_p ds}$. This definition is proportional to the total amount of information gathered $\int_p h(s) ds$, and is inversely proportional with the travel distance $\int_p ds$. When the path is closed, the region $\mathbb{O}$ inside $p$ can be segmented by the tracing path in $\mathbb{T}$, and thus labeling the region $\mathbb{O}$ as occupied.  This is equivalent to an exploration rate of $\frac{\int_p h(s) ds + \int_{l} h(s) ds}{\int_p ds}$, where $l$ is the shortest coverage path inside region $\mathbb{O}$. This shows analytically that the exploration rate of the latter case is greater than the former case. Thereby enclosing the contour tracing path brings a higher exploration rate.  $\blacksquare$

Without loss of generality, the contour tracing can be accomplished in multiple ways. For instance, a robot can sweep a force sensor along the object's perimeter \cite{ahmad1990shape}. If equipped with a multi-taxel tactile sensor, the contour tracing can be accomplished by pressing on an edge and simultaneously following the edge direction \cite{lepora2019pixels}. Such approaches require surfaces to be continuous and uniform enough in order to slide the sensor smoothly. In contrast, many real objects have sharp edges or corners that do not comply with the smoothness condition, which is not consistent with the basic assumption of the previous techniques.
To address this problem, reactive rhythmic spiral movement patterns generated by Hopf bifurcation \cite{hassard1981theory} are introduced. This is denoted as the Hopf oscillator (introduced later in Sec.~\ref{met:osimp}), which has been applied to planning versatile robotic locomotions, such as swimming \cite{hu2014parameter}, hopping \cite{buchli2005dynamical}, and quadrupled walking gaits \cite{liu2017hopf}.

\subsection{Object Searching (OS)} \label{met:os}

{\revcolor{red}In this section, methods are proposed for the integration of the contour tracing into the informative path planning framework for achieving tactile exploration in a scene.} Let us define the \textit{search space} as the subspace of $\mathbb{T}$ where object searching is performed. At time step 0, $\mathbb{S}^0 = \mathbb{T}$. If at time step $t$, objects occupying regions $\mathbb{O}^1, \mathbb{O}^2, \cdots, \mathbb{O}^k$ have been discovered, the search space $\mathbb{S}^t$ at time $t$ can be obtained by removing those occupied regions from the task space $\mathbb{T}$ i.e., $\mathbb{S}^t = \mathbb{T} - \Sigma_{i=1}^{k}\mathbb{O}^i$, where ``$-$'' and ``$\sum$'' operators are defined as the \textit{difference} and \textit{union} operation between sets, respectively. Following the definition of the search space, we can articulate remark 1.

\textbf{Remark 1}: Searching objects in $\mathbb{S}^t$ can be simplified as a pure exploration problem. 

Since the search space $\mathbb{S}^t$ does not include any observed occupied regions, there are no historical observations acting as priors for the prediction of the object locations to be discovered.
Besides, when assuming the locations of objects are independent of each other, it is also not possible to use discovered objects as priors. As a result, the only feasible policy for searching objects in $\mathbb{S}^t$ is to collect information from unexplored regions, which is subject to the definition of the pure exploration problem.

Next, let us define the occupancy function $f$, which is given in Eq.~(\ref{eq:occupancyfunc}). Gathering observations from different locations builds up the observation set $\chi$. We assume the occupancy state of a position $\mathbf{x}_o$ can be acquired from a tactile sensor in the proposed implementation.

\begin{equation}
\left\{  
    \begin{array}{l}
    f(\mathbf{x}_o) = 1 \;\;\text{if position }\;\mathbf{x}_o\;\text{is occupied}\\
    f(\mathbf{x}_o) = 0 \;\;\text{if position }\;\mathbf{x}_o\;\text{is not occupied}
    \end{array}
\right.
\label{eq:occupancyfunc}
\end{equation}

The acquisition function $h(\textbf{x})$ is defined as Eq.~(\ref{eq:utility}).

\begin{equation}
    h(\textbf{x}) = \sigma(\textbf{x})
    \label{eq:utility}
\end{equation}

In Eq.~(\ref{eq:utility}), $\sigma(\textbf{x})$ is the standard deviation of the estimated occupancy function $\hat{f}$. When ignoring the transition path, the sampling process is subject to the \textit{uncertainty sampling} convention in the Bayesian optimization setting \cite{blanchard2020informative}. Note that $h(\textbf{x})$ evolves as the sampling proceeds.

We leverage Gaussian Process regression (GPR) to estimate $\sigma (\textbf{x})$. This estimator is given by Eq.~(\ref{eq:GP}), where $\kappa$ is a kernel function. In our setting, we use radial basis function (RBF) as the kernel function. $\mathbf{x}_* \in \mathbb{R}^2$ is the position to be queried for $\sigma (\textbf{x}_*)$.

\begin{equation}
\sigma^{2}\left(\mathbf{x}_{*}\right) = \kappa(\mathbf{x}_*, \mathbf{x}_*)-\mathbf{k}^T_{*}\left(\mathbf{K}+\sigma_n^{2} \mathbf{I}\right)^{-1} \mathbf{k}_*
\label{eq:GP}
\end{equation}

$\mathbf{k}_{*}$ and $\mathbf{K}$ are defined in Eq.~(\ref{eq:GP_a}) and Eq.~(\ref{eq:GP_b}), respectively:

\begin{equation}
\mathbf{k}_{*} = \left[ \kappa(\mathbf{x}_*, \mathbf{x}_o^{1}), \kappa(\mathbf{x}_*, \mathbf{x}_o^{2}),...,\kappa(\mathbf{x}_*, \mathbf{x}_o^{N}) \right]
\label{eq:GP_a}
\end{equation}

\begin{equation}
\mathbf{K} = \begin{bmatrix}
\kappa(\mathbf{x}_o^1, \mathbf{x}_o^1) & \kappa(\mathbf{x}_o^1, \mathbf{x}_o^2)  & \cdots   & \kappa(\mathbf{x}_o^1, \mathbf{x}_o^N)  \\
\kappa(\mathbf{x}_o^2, \mathbf{x}_o^1) & \kappa(\mathbf{x}_o^2, \mathbf{x}_o^2)  & \cdots   & \kappa(\mathbf{x}_o^2, \mathbf{x}_o^N)  \\
\vdots & \vdots  & \ddots   & \vdots  \\
\kappa(\mathbf{x}_o^N, \mathbf{x}_o^1) & \kappa(\mathbf{x}_o^N, \mathbf{x}_o^2)  & \cdots   & \kappa(\mathbf{x}_o^N, \mathbf{x}_o^N)  \\
\end{bmatrix}
\label{eq:GP_b}
\end{equation}

The goal of the object searching task is to bring to zero the unexplored regions in the search space as the number of samples reaches infinity, so that an object can be eventually detected regardless of its size. We show this can be achieved by following proposition 2.

\textbf{Proposition 2}: Complete coverage of the obstacle-free regions in the task space $\mathbb{T}$ can be achieved by the following hybrid policy: 

1) \textit{object searching policy}: search objects by walking towards $\mathbf{x}_{max}^{t} = \text{max}_{\textbf{x} \in  \mathbb{S}^{t}} h^{t}(\textbf{x})$. When $\mathbf{x}_{max}^t$ is reached at time $t+\beta$, continue to search object by replanning a new target $\mathbf{x}_{max}^{t+\beta} = \text{max}_{\textbf{x} \in  \mathbb{S}^{t+\beta}} h^{t+\beta}(\textbf{x})$, and then iterate the above procedures. 

2) \textit{contour tracing policy}: apply contour tracing immediately when an object is encountered at time $t+\gamma$ before reaching $\mathbf{x}_{max}^t$. The object should be fully enclosed into $\mathbb{O}^i$. After that, resume to the object searching policy i.e., to follow a path that is towards a new target $\mathbf{x}_{max}^{t+\gamma} = \text{max}_{\textbf{x} \in  \mathbb{S}^{t+\gamma}} h^{t+\gamma}(\textbf{x})$, where $\mathbb{S}^{t+\gamma} = \mathbb{S}^t - \mathbb{O}^i$.

\textbf{Proof}: The proof can be completed by analyzing on two sub-policy cases, respectively:

\begin{enumerate}
    \item For the object searching policy, $\mathbf{x}_{max}^t$ is reached at time step $t+\beta$, Then the robot will continue moving to the next planned target $\mathbf{x}_{max}^{t+\beta}$. Repeating this target chasing approach infinitely will lead to complete coverage of the obstacle-free area in the task space. This is the space-filling property of the Maximum Squared Error (MSE) sampling, which is referred to Theorem 6 and Theorem 7 in literature \cite{vazquez2010convergence}.  
    \item The contour tracing policy starts to execute at the same time when a contact event occurs, by which the object searching policy is interrupted before reaching $\mathbf{x}_{max}^t$. Since an occupied region will be found, this will result in a reduced search space (remark 1). Given the fact that there is a finite number of objects in the task space ($N_o$ objects in total), transition to contour tracing policy will occur $N_o$ times at most. After that, no objects will be in the search space and maximum task space coverage will be attained by the object searching policy.  $\blacksquare$
\end{enumerate}

\subsection{Policy Design} \label{met:osimp}

Proposition 2 defined the basic framework of the algorithm proposed. {\revcolor{red}We implemented this algorithm by a finite state machine with two states: OS and CT. The robot starts in OS state to search objects until a contact event occurs. Then, the robot switches to CT to acquire contact points by tracing the object contour. Once the contour tracing path is closed, the state machine transfers back to OS to search for the next object. The overall process is detailed in Alg.~\ref{alg:statemachine}.}

\subsubsection{{\revcolor{red}Policy for OS state}}

\begin{figure}[tb]
    \centering
    \subfloat[random sampling]{\includegraphics[width=0.46\linewidth,trim=40 40 40 40,clip]{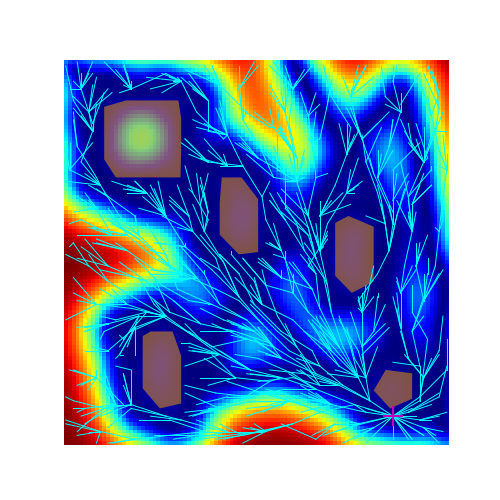}}
    \subfloat[heuristic sampling]{\includegraphics[width=0.46\linewidth,trim=40 40 40 40,clip]{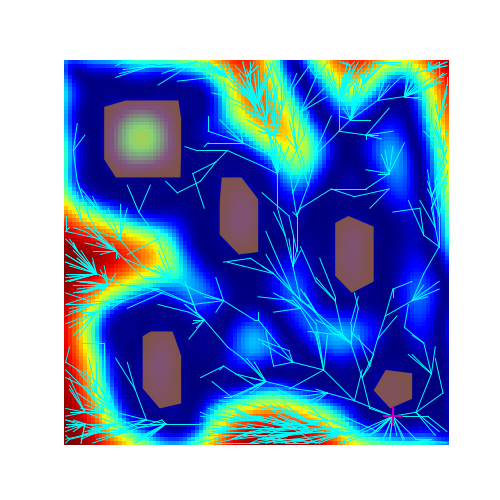}}
    \caption{Comparison of different sampling methods used in tree expansion. Red regions have higher uncertainty than blue regions. Orange polygons are the estimated objects. Heuristic sampling guides the tree expansion towards high uncertainty regions, and therefore, has a higher chance of finding paths with more information than random sampling.}
    \label{fig:sampling_compare}
\end{figure}

The object searching procedure target to plan an informative path based on the information acquisition function $h(\mathbf{x})$ and an occupancy map. Here $h(\mathbf{x})=\sigma(\mathbf{x})$ is obtained by calculating Eq.~(\ref{eq:GP}). Overall, the proposed informative path planner expands the idea from a sampling-based IPP framework \cite{schmid2020efficient}. A Rapidly-exploring random tree \cite{lavalle1998rapidly} is expanded inside the search space. The proposed algorithm is coined as Tactile object searching (TOS) planner. The pseudo-code of TOS is given in Alg.~\ref{alg:os_rrt}. A tree structure is maintained by a vertex set $V$ and an edge set $E$. The tree is expanded in the same way of RRT$^{*}$, in which the tree branch steers towards a stochastic target $\mathbf{x}_{sn}$ from its nearest node $\mathbf{x}_{near}$ \cite{karaman2011sampling}. Unlike canonical RRT$^{*}$ that aims to reach a single target, a heuristic sampling process is used to guide the tree expansion towards regions with high values in $h(\mathbf{x})$. To generate each tree node $\mathbf{x}_{new}$, a total number of $\mathit{n}$ candidate samples in the search space $\mathbb{S}$ are generated uniformly. The $\mathbf{x}_{sn}$ is chosen by roulette selection (Eq.~(\ref{eq:roulette}), with $k_r=3$) from $n$ candidate points. Fig.~\ref{fig:sampling_compare} gives a comparison between exploration trees from heuristic sampling and the canonical uniform sampling using the same number of samples.

\begin{equation}
p(\mathbf{x}_i)=\frac{\sigma(\mathbf{x}_i)^{k_r}+\epsilon}{\Sigma_{j=1}^{n} (\sigma(\mathbf{x}_i)_{j}^{k_r} + \epsilon)} 
\label{eq:roulette}
\end{equation}

Once the tree has been expanded, $\mathbf{x}_{max}^t = \text{max}_{\textbf{x} \in \mathbb{S}^t} h^t(\textbf{x})$ can be approximated by using vertices in the vertex set $V$ (i.e., $\hat{\mathbf{x}}_{max}^t = \text{max}_{\textbf{x} \in V} h^t(\textbf{x})$). The planned path can be obtained by backtracking from $\hat{\mathbf{x}}^t_{max}$ to the robot position $\mathbf{x}_{robot}$ (tree root).

\begin{algorithm}[tb]
    \SetKwInOut{Input}{Input} 
    \Input{ $\mathbb{T}$: task space \\}
    \KwResult{ GP: a Gaussian Process model of the scene}
    
   \SetAlgoLined\DontPrintSemicolon
   initialize(GP), state = ``OS''\\
   $\chi = \varnothing$\\
   \While{not terminated}{
        \If{state is ``OS''}{
            path = TOS\_planner(GP, $\chi$)\\
            \While {path not empty} {
                $\mathbf{x}_{OS}$ = pop(path)\\
                $S$ = move\_and\_sensor\_observe($\mathbf{x}_{OS}$)\\
                $\chi = \chi \cup (\mathbf{x}_{OS}, S)$\\
                \If{$S$ is in contact state}{
                    state = ``CT'', break\\
                }  
            }
        }
        \If{state is ``CT''}{
            \While{contour not closed}{
                $\mathbf{x}_{CT}$ = Hopf\_oscillator($S$)\\
                $S$ = move\_and\_sensor\_observe($\mathbf{x}_{CT}$)\\
                $\chi = \chi \cup (\mathbf{x}_{CT}, S)$\\
            }
            state = ``OS''\\
        }
        update(GP, $\chi$)

   }

\caption{Hybrid Tactile Exploration Policy}
\label{alg:statemachine}
\end{algorithm}

\begin{algorithm}[htb]
    \SetKwInOut{Input}{Input} 
    \Input{ 
    $\mathbb{T}$: Task space\\
     $\chi$: Sensor observation set\\
    GP: Gaussian Process occupancy function}
       
    \KwResult{ $P$: A trajectory as a array of points }
    
    \SetAlgoLined\DontPrintSemicolon
    
    \SetKwFunction{algo}{object\_search}
    \SetKwProg{myalg}{Algorithm}{}{}
    \myalg{TOS\_planner(GP, $\mathbb{T}$)}{
        $V = \{\mathbf{x}_{robot}\}, E = \varnothing, P = \varnothing$\\
        $\mathbb{O}$ = predict\_occupancy\_polygons($\chi$)\\
        \For{$i = 1, 2, \cdots, N_{tree}$}{
            $\mathbf{x}_{sn}, \sigma(\mathbf{x}_{sn})$ = uncertainty\_sampling($\mathbb{T}$, $\mathbb{O}$)\\
            $\mathbf{x}_{near} = near(V, \mathbf{x}_{sn})$\\
            $\mathbf{x}_{new} = steer(\mathbf{x}_{near}, \mathbf{x}_{sn}, d_{near})$\\
            \If{collision\_free($\mathbf{x}_{new}$,$\mathbf{x}_{near}$)}{
                $V = V\cup \mathbf{x}_{new}$
            
                $ \chi_{near}$ = query\_near\_set($\mathbf{x}_{new}$)\\
                \For{$\mathbf{x}_{near}^*\in \chi_{near}$}{
                    \If{$||\mathbf{x}_{near}^*-\mathbf{x}_{new}||_2$ + dis\_root($\mathbf{x}_{new}$)\textless dis\_root($\mathbf{x}_{near}^*$)}{
                        set\_parent($\mathbf{x}_{near}^*$, $\mathbf{x}_{new}$, $E$)
                    }
                }
            }
        }
        $\mathbf{x}_{max} = \text{max}_{\mathbf{x}_v \in V}(\sigma(\mathbf{x}_{v})), \mathbf{x}_{it} = \mathbf{x}_{max}, P = P \cup \mathbf{x}_{it}$\\
        \While{parent($\mathbf{x}_{it}$, $E$, $V$) $\neq \varnothing$}{
            $\mathbf{x}_{it} = \text{parent}(\mathbf{x}_{it}, E, V)$\\
            $P = P \cup \mathbf{x}_{it}$
        }
        \KwRet $P$
    }
    
    \SetKwRepeat{Do}{do}{while}

    \SetKwProg{Fn}{Function}{:}{}
    \Fn{$uncertainty\_sampling(\mathbb{T}, \mathbb{O})$}{

    $\mathbb{S} = \mathbb{T} - \mathbb{O}$\\
    $\mathbf{X}_{\mathbb{S}}$ = uniform\_sampling($\mathbb{S}$)\\
    
    $\sigma(\mathbf{X}_{\mathbb{S}})$ = GP($\mathbf{X}_{\mathbb{S}}$)\\
    $\mathbf{x}_{s}$ = roulette\_select($\mathbf{X}_{\mathbb{S}}$, $\sigma(\mathbf{X}_{\mathbb{S}})$)\\
    \KwRet $\mathbf{x}_{s}$, $\sigma(\mathbf{x}_{s})$
    }

\caption{TOS: Tactile Object Searching}
\label{alg:os_rrt}
\end{algorithm}

\subsubsection{{\revcolor{red}Policy for CT state}}

Hopf oscillator is used to generate a forward propagation path along the object contour. For a given central point $\mathbf{x}_c^{i} = (x_c^i, y_c^i)$, The dynamic equation of the Hopf oscillator in a Cartesian coordinate is described as Eq.~(\ref{eq:hopf}).

\begin{equation}
\begin{array}{l}
\hat{x} = x_s - x_c^i\\
\hat{y} = y_s - y_c^i\\
\dot{\hat{x}}=\gamma\left(r_H^2-\hat{x}^2-\hat{y}^2\right) \hat{x}-2 \pi f_H \hat{y} \\
\dot{\hat{y}}=\gamma\left(r_H^2-\hat{x}^2-\hat{y}^2\right) \hat{y}+2 \pi f_H \hat{x}
\label{eq:hopf}
\end{array}
\end{equation}

Where $\mathbf{x}_s = (x_s, y_s)$ is the sensor's position in Cartesian coordinates. $f_H$ is the oscillatory frequency. {\revcolor{red}$\gamma$ is a gain parameter which regulates the spiral motion to be within a circle of radius $r_H$.} When a contact event occurs, the algorithm changes the tooltip (where the sensor is mounted on) velocity direction, in such a way that it causes a ``bouncing off'' effect. 
This is implemented by updating the central point from $\mathbf{x}_c^i$ to $\mathbf{x}_c^{i+1}$ according to Eq.~(\ref{eq:hopf_contact1}) {\revcolor{red}for a binary touch sensor (even for a single whisker).}
\begin{equation}
\begin{array}{l}
    x_c^{i+1} = 2x - x_c^{i}\\
    y_c^{i+1} = 2y - y_c^{i}
\end{array}
\label{eq:hopf_contact1}
\end{equation}

Alternatively, if the planar contact normal vector $\mathbf{c} \in \mathbb{R}^2$ is observable, the central point can also be calculated by Eq.~(\ref{eq:hopf_contact2}). {\revcolor{red}The benefit of having the contact normal is to place the next spiral center $\mathbf{x}_{c}^{i+1}$ to be along the surface direction, aiming to achieve a step size of approximate $2 r_H$. This in turn reduces the variance in the interval distances between contact points. In our experiment, the contact normal vector is estimated to be along the normal vector of the ``ring'' on the location of the contact point. The estimation accuracy depends on the density of whiskers, which can be scaled up as a design choice.}

\begin{equation}
\begin{array}{l}
    \mathbf{\mathbf{w}}=[0, 0, 1]^{T} \times \mathbf{c} = [\mathbf{w}_{12}, 0]^{T}\\
    \mathbf{x}_{c}^{i+1} = \mathbf{x}_s + \frac{r_{H}\mathbf{w}_{12}}{|\mathbf{w}_{12}|}
\end{array}
\label{eq:hopf_contact2}
\end{equation}

\begin{figure}[tb]
    \centering
    \subfloat[]{\includegraphics[width=0.48\linewidth,trim=60 5 80 20,clip]{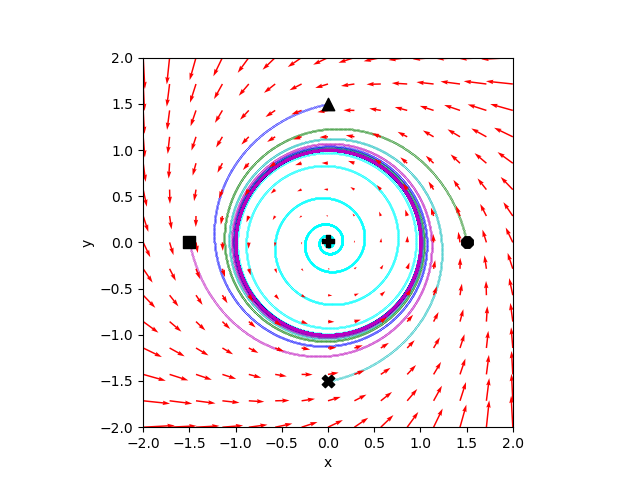}}
    \subfloat[]{\includegraphics[width=0.48\linewidth]{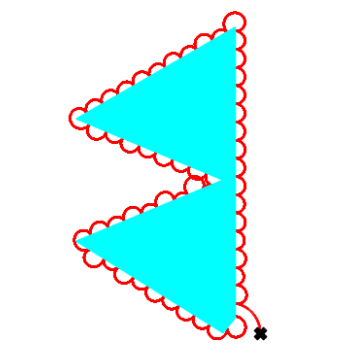}}
    \caption{ Demonstrations of the spiral movement pattern of the Hopf oscillator. (a) A plot of the phase portrait (red arrows), and the trajectories starting from 5 different initial positions (marked as \ding{108}\ding{54}\ding{115}\ding{110}\ding{58}, respectively). (b) An example trajectory (red) of using Hopf oscillator to trace the contour of one non-convex object (cyan polygon). }
    \label{fig:hopf_illustration}
\end{figure}

{\revcolor{red}
By transiting the position of the spiral center, the dynamics in Eq.~(\ref{eq:hopf}) will result in contact break and a trajectory which moves the sensor forward. An example of the described motion is shown in  Fig.~\ref{fig:hopf_illustration}.}

\subsection{Object Contour Extraction} \label{met:ol}
The object contour can be estimated by directly connecting the contact points according to the contact sequence found through the Hopf oscillator. While this is a simple technique, it brings clear advantages when compared to conventional point cloud segmentation approaches using clustering (e.g., mean-shift clustering \cite{kaboli2019tactile}). {\revcolor{red}First, the object and point segmentation (attributing points to objects) can be trivially achieved, because each transition from OS to CT only introduces exact one new object, which all the following contact samples belong to. This is in contrast to the clustering technique which processes all objects inside the scene at once, for which determining the number of clusters, their locations, and point labeling are all treated as independent approximation steps that lack robust solutions.} It has been reported that the performance of clustering techniques is degraded when objects are spatially entangled, or when the point cloud density is unevenly distributed \cite{wiwie2015comparing}. In comparison, segmentation by contour tracing is unlikely to be affected by these factors, as shown by the results in Sec.~\ref{localization_acc}. Second, an algorithm that is widely used to calculate the concave hull polygon from a point cluster is the $\alpha$-shape \cite{edelsbrunner1983shape}, which involves higher computational cost, and an ad-hoc parameter $\alpha$ that requires manual efforts in tuning. In contrast, our introduced method is computationally cheap and parameter-free. {\revcolor{red}Alternatively, after segmenting the points as contours by our approach, it is possible to smooth the contour by other geometric processing techniques, such as Gaussian process implicit surface \cite{williams2006gaussian} demonstrated in Fig.~\ref{fig:real_contours}.}

\section{Object Recognition} \label{sec:recog}

\subsection{Problem Definition} \label{sec:res_or_problem_def}

The key idea presented in this section relies on the fact the contour shapes convey categorical information. Such information can enable object classification among a wide category of objects when used appropriately. For this purpose, a deep learning classifier is used for object classification using object contour.  In addition, the discriminability of the proposed classifier is improved by incorporating very few probing points in $\mathbb{R}^3$ space.

The problem formulation of classifying an object using contour points can be expressed as: \textit{finding function $F$ that predicts the object label $y_{obj}=F(P_{ct}, P_{fs})$}. $P_{ct}$ is a point set that belongs to an object contour $C_i \in \mathbb{R}^2$. $P_{fs}$ is a point set in $\mathbb{R}^3$ that is within the object surface and is reachable by the end-effector. $P_{ct}$ can be obtained by contour tracing (Sec.~\ref{sec:ct}). $P_{fs}$ is collected during Feature Sampling (FS) procedure described in Sec.~\ref{sec:hybrid}. While the method for obtaining $P_{fs}$ is task-dependent, adding a few discrete contacts inside the estimated contour polygon $\hat{C}_i$ from above of the object provides a general approach. {\revcolor{red}This allows to generalize the object characterization from a planar space to a spatial region.} Given the number of total contacts $K_{fs}$, and planar locations $(x_{fs}^j, y_{fs}^j), j\in 1, 2, \cdots, K_{fs}$ that are inside the contour polygon $\hat{C}_i \in \mathbb{R}^2$, the end-effector is lowered at each of those locations from above of the object, until the sensor reaches the object at $z=z_{fs}^j$. {\revcolor{red}If no contact is detected, the $z_{fs}^j$ is set to zero, indicating that the object is hollow at that point. This allows the algorithm to work on versatile objects without assuming priors about the object's shape and convexity.} Note that the position of the first contact $(x_{fs}^0, y_{fs}^0)$ for each object is always chosen as the centroid point of the object's estimated contour polygon $\hat{C}_i$.

\subsection{Design of classifier} \label{sec:classifier}

\begin{figure}
    \centering
    \includegraphics[width=\linewidth]{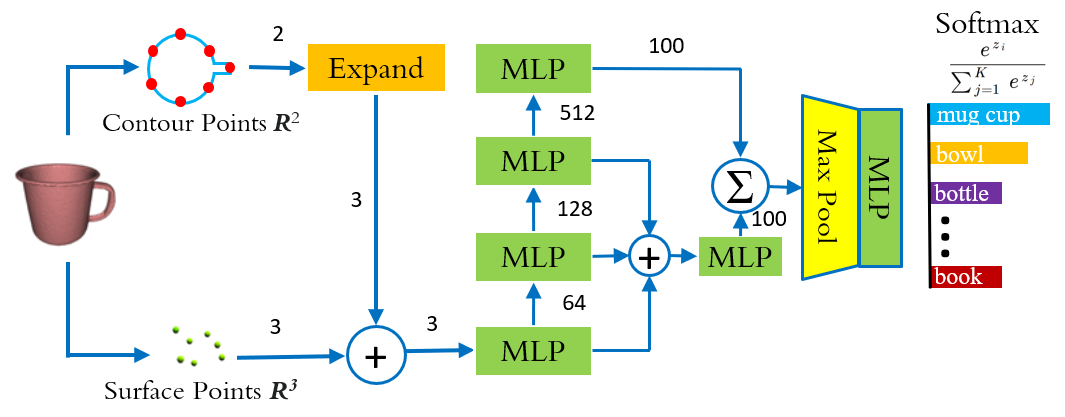}
    \caption{Schematic of the proposed classifier based on the planar contour points and volumetric surface points.}
    \label{fig:classifier}
\end{figure}

Point clouds have been classified by deep neural networks \cite{luo2017robotic, qi2017pointnet}. We will leverage this to design a classifier that uses both contour points and probing points as the input observations. The proposed network is named as CT-Net. The network architecture is described in Fig.~\ref{fig:classifier}. In this network, contour points in $P_{ct} \in \mathbb{R}^2$ are transformed into $\mathbb{R}^3$ by increasing one dimension (appending a zero), and then being concatenated with probing points in $P_{fs} \in \mathbb{R}^3$. Then, each point is transformed to higher dimensions progressively using Multi-Layer Perception (MLP) with ReLU activation function. Side branches are added, to alleviate the gradient vanishing problem by creating shortcuts for intermediate layers. This is accomplished by concatenating all feature branches first (``+'' operator in figure), and then reducing the feature dimension by using another MLP. The transformed intermediate features are fused into the last MLP layer's output by a sum operation (``$\sum$'' operator in figure). 

Since the point sequence can vary (even for planar contours, since the sampling does not always start from the same point), an aggregation function is used to obtain the invariance from the point order. To achieve such invariance, a max aggregation \cite{qi2017pointnet} is used to calculate the object latent feature. Next, a canonical MLP classifier with a softmax function is used to calculate the categorical possibility vector.

\section{Results on Whisker Sensing}\label{sec:res_whisker}

\begin{figure}[htb]
    \centering
    \includegraphics[width=0.9\linewidth]{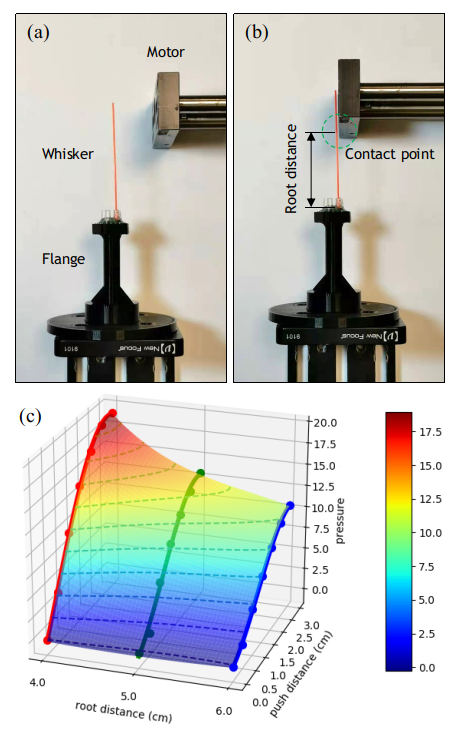}

    \caption{(a) The motor testbed that pushes the whisker horizontally. (b) The motor is in the critical contact state with the whisker. (c) Sensor's pressure reading as a function of root distance, and push distance. }
    \label{fig:surface_curve}
\end{figure}

The performance of the whisker sensors was characterized by pushing the whiskers under different conditions. Linear motor (LinMot PS01-23 × 80) was used to supply a linear reciprocating motion at a speed of 1 m/s. The PCB was mounted on a holder flange and fixed onto a base plate. The fixing pose assures the whisker to be perpendicular to the moving direction of motor's pusher, as shown in Fig.~\ref{fig:surface_curve} (a) and (b). Initially, the motor and the whisker were aligned to be in the critical contact state (zero air gap with no pressure offset observed from the barometer). The pushing distance of the motor was then set as 0 mm, 5 mm, 10 mm, 15 mm, 20 mm, 25 mm, and 30 mm, respectively.  It could also be observed that the sensor's output is a function of the ``root distance'', which is defined as the distance between the contact point and the barometer in the critical contact state.  The root distance was set as 60 mm, 50 mm and 40 mm, respectively. This results in a combination of $7\times3=21$ experiment configurations.

The resultant pressures from barometer sensors are shown in Fig.~\ref{fig:surface_curve} (c). B-spline interpolation was used to obtain the intermediate value (surface) based on the sensor's raw pressure outputs (dots).  {\revcolor{red}First, when the root distance is fixed, the pressure reading increases monotonously as the pushing distance increases. This monotonicity implies that when the robot can actively determine the pushing distance (e.g., horizontally offsetting the tool from the initial contact position),  it is theoretically possible to calculate the root distance as a function of the sensor's pressure reading, by which the contact position can be estimated.} But the sensitivity may decrease if the pushing distance were too large, as the slope of the pressure gradually decreases with the pushing distance. Second, under the same pushing distance, the pressure reading decreases when the root distance increases (farther from the barometer). This implies that there may exist a maximum length limit when choosing the whisker filaments.

\begin{figure}[htb]
\includegraphics[width=0.98\linewidth]{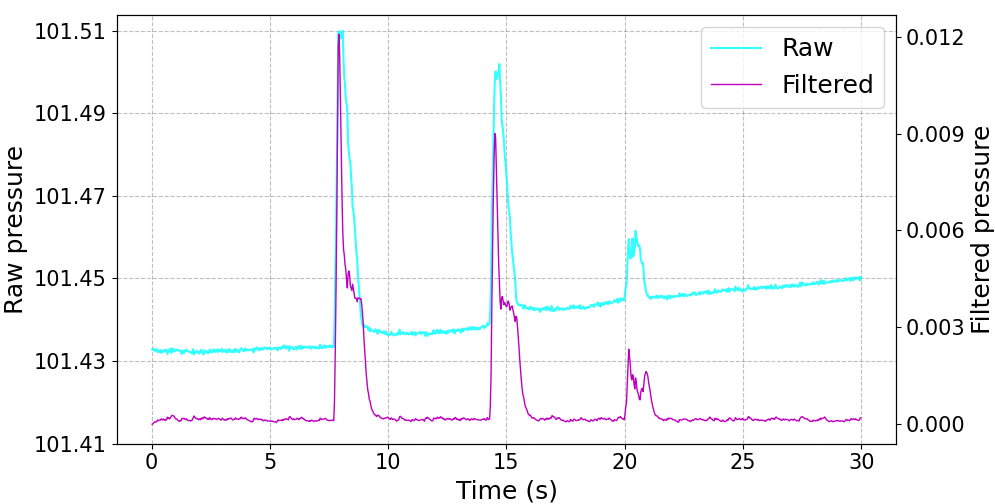}
\caption{Sensor's reading in the time domain. Cyan curve is the raw data from the sensor. Magenta curve is obtained by processing the raw data with filters. } 
    \label{fig:real_data}
\end{figure}

Fig.~\ref{fig:real_data} shows the sensor's output in the time domain. The experiment was conducted with a real robot controlled by the Hopf oscillator.  Overall, there are three contact events that correspond to three signal peaks. Besides, the raw pressure data from the barometer's  reading drifted by 0.017 kPa in a time period of 30 seconds, from 101.433 kPa to 101.450 kPa. Nevertheless, the drifting effect can be mitigated by the filter module introduced in Sec.~\ref{sec:wsd}, by which the magenta curve with a higher signal-noise ratio was obtained.

\section{Results on Tactile Exploration}\label{sec:res_te}

\subsection{Experiment Configuration}

The proposed tactile spatial exploration algorithm was evaluated both in simulation, and with a real robot. A UR16e robot was equipped with the developed whisker sensor. The algorithm parameters used in both evaluation settings are shown in Table.~\ref{tbl:seg}. In particular, the evaluations focused on the following: 1) the efficiency of exploration, and 2) the quality of the object  shape recovered from the tactile observations.

\begin{table}[htb]
\centering
\vspace{-3pt}
\caption{parameters used in the experiments.}
\vspace{-3pt}
\label{tbl:seg}
\begin{tabular}{cccc}
\hline\hline

Symbol & Description                  &  Simulation  & Real \\\hline 
$W_S$ & Scene width (m) & 1.0 & 0.59\\
$H_S$ & Scene length (m)  & 1.0 & 0.62\\
$r_{H}$ & Oscillator radius (m)  & 0.025 & 0.02\\
$\gamma$ & Oscillator radial gain & 10 & 10\\
$f_H$ & Oscillator frequency (Hz) & 0.5 & 0.5\\
$N_{tree}$&No. of nodes in tree expansion & 1000 & 1000\\
$d_{near}$&neighborhood distance threshold (m) & 0.10 & 0.10\\ 
$l_{RBF}$ & RBF kernel lengthscale (m)  & 0.08 & 0.08\\
$\sigma_n$ & GP noise level & 0.02 & 0.02\\

\hline\hline
\end{tabular}
\vspace{-5pt}
\end{table}

\subsection{Exploration Efficiency} \label{sec:exp_eff}

The exploration efficiency can be quantified using two metrics: 1) \textit{Scene Uncertainty} measures the acquisition function $h(\mathbf{x})$ averaged over all positions in the task space (Eq.~(\ref{eq:metric1})). {\revcolor{red}Lower values indicate the environment have been better explored, resulting in less uncertainty remained.}
\begin{equation}
    U^{S} = \frac{\iint_{\mathbb{T}} \sigma(\mathbf{x}) dS}{\iint_{\mathbb{T}} dS}
    \label{eq:metric1}
\end{equation}
2) \textit{Contour Uncertainty} measures the acquisition function $h(\mathbf{x})$ averaged over all positions within the object contour $C_i$  (Eq.~(\ref{eq:metric2})). {\revcolor{red}Lower values are better, as they indicate less uncertainty remained on the object's contour.}

\begin{equation}
    U^{C} = \frac{\sum_{i=1}^{N_o} \int_{C_i} \sigma(\mathbf{x}) ds}{\sum_{i=1}^{N_o} \int_{C_i} ds}
    \label{eq:metric2}
\end{equation}

The proposed algorithm was compared both qualitatively and quantitatively with two baselines. The first baseline is the method proposed by Kaboli et al. (2019) \cite{kaboli2019tactile}. {\revcolor{red}This chosen baseline is known to be the state-of-the-art for leveraging tactile sensing for scene exploration, and can be adopted to our experiment setting without further modifications.} In this work, the sensor collects contact samples by progressively selecting line paths that are parallel to the $x$ and $y$ axis. The line parameters are chosen according to a constraint, which is specified to make the line crosses $\text{max}_{\textbf{x} \in \mathbb{T}}  \sigma^2(\textbf{x})$, aiming to increase the information acquisition. We denote this method as ``Line Sweep'' due to its motion pattern. Second, another baseline is the TOS planner alone, which is a vanilla object searching policy without incorporating the contour tracing (also denoted as ``pure object searching policy''). This comparative experiment can reveal the contribution of contour tracing in object characterization.

\captionsetup[subfigure]{labelformat=empty,aboveskip=-3pt}
\begin{figure}[htb]
    \centering
    \vspace{-3mm}
    \subfloat[\textcircled{1}]{\includegraphics[width=0.33\linewidth,trim=30mm 10mm 30mm 10mm, clip=true]{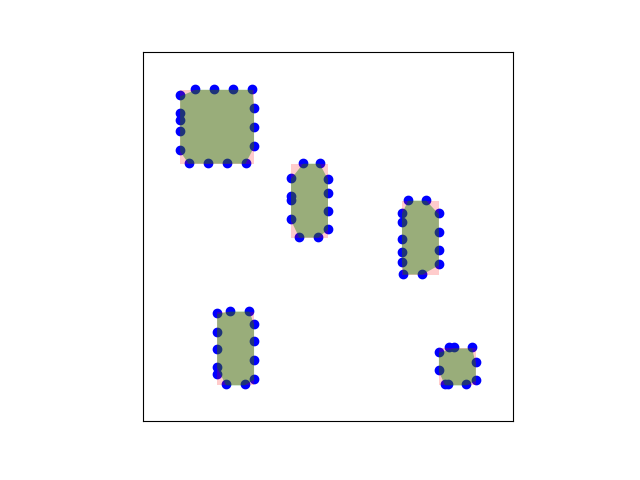}}
    \subfloat[\textcircled{2}]{\includegraphics[width=0.33\linewidth,trim=30mm 10mm 30mm 10mm, clip=true]{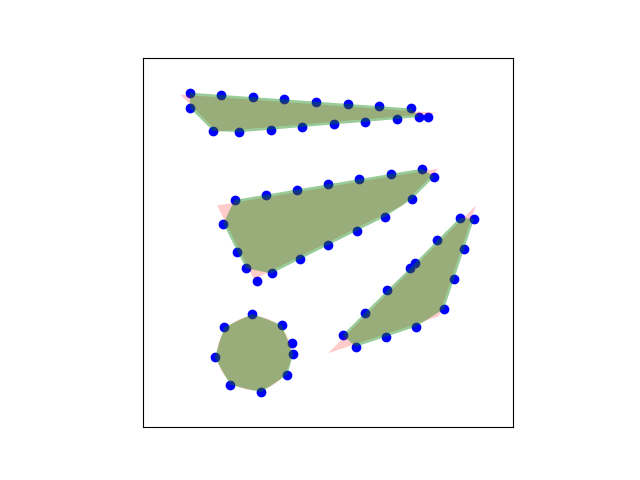}}
    \subfloat[\textcircled{3}]{\includegraphics[width=0.33\linewidth,trim=30mm 10mm 30mm 10mm, clip=true]{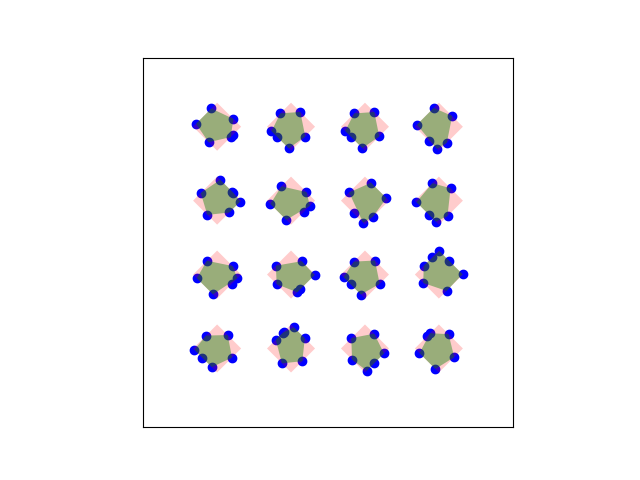}}
    
    \subfloat[\textcircled{4}]{\includegraphics[width=0.33\linewidth,trim=30mm 10mm 30mm 10mm, clip=true]{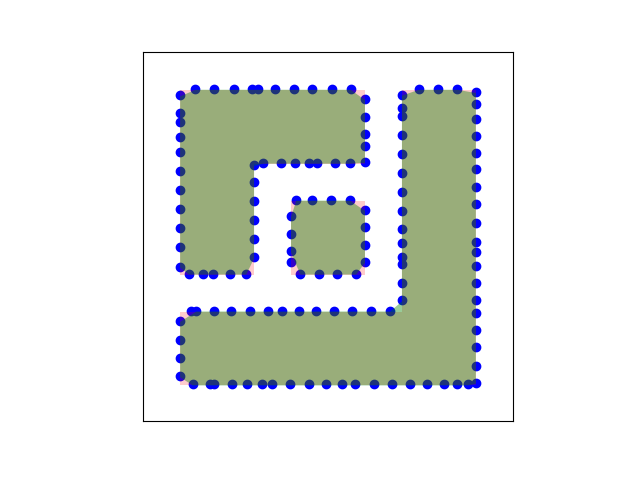}}
    \subfloat[\textcircled{5}]{\includegraphics[width=0.33\linewidth,trim=30mm 10mm 30mm 10mm, clip=true]{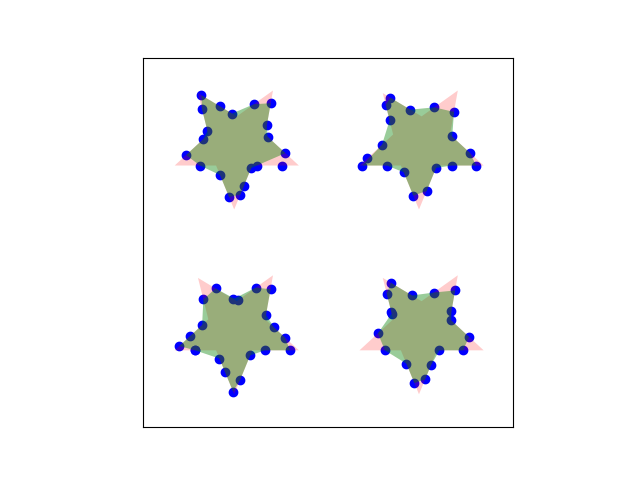}}
    \subfloat[\textcircled{6}]{\includegraphics[width=0.33\linewidth,trim=30mm 10mm 30mm 10mm, clip=true]{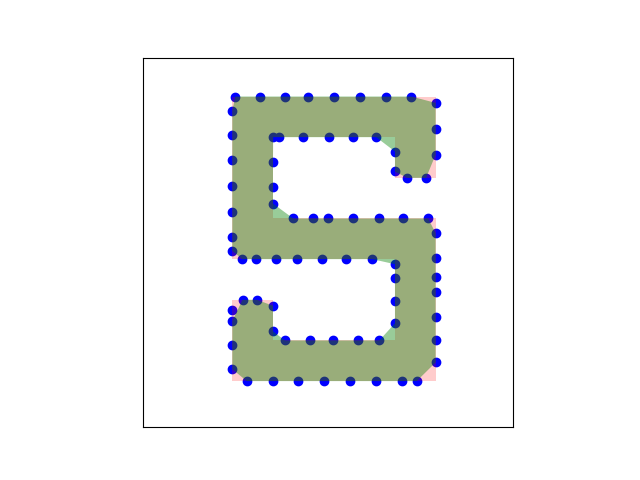}}
    
    \caption{Reconstructed objects from contact points (green region) versus the ground truth object (pink region, and dark green when overlapped with the reconstruction). Dots are the sampled contact points. \textcircled{1} and \textcircled{2} are randomly distributed primitives, \textcircled{3} small objects, \textcircled{4} inter-occluded objects, \textcircled{5} pentagrams, \textcircled{6} a S-shape object.}
    \label{fig:reconstruct}
\end{figure}
\captionsetup[subfigure]{labelformat=parens,aboveskip=0pt}

\begin{figure}[htb]
    \centering
    \subfloat[Scene \textcircled{1}, $U^S$]{\includegraphics[width=0.48\linewidth]{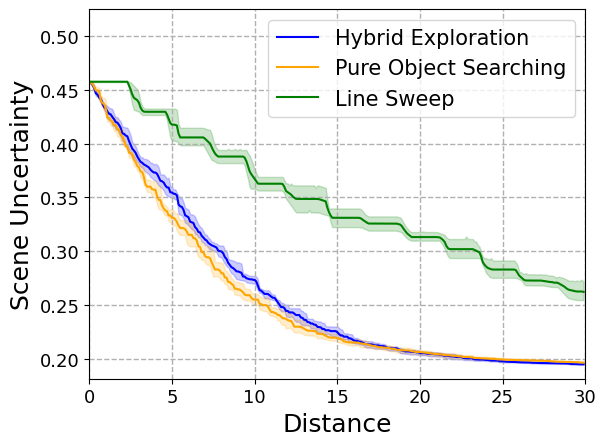}}
    \subfloat[Scene \textcircled{1}, $U^C$]{\includegraphics[width=0.48\linewidth]{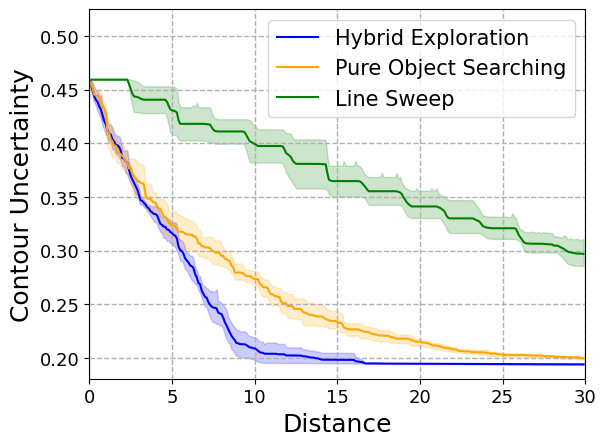}}
    
    \subfloat[Scene \textcircled{4}, $U^S$]{\includegraphics[width=0.48\linewidth]{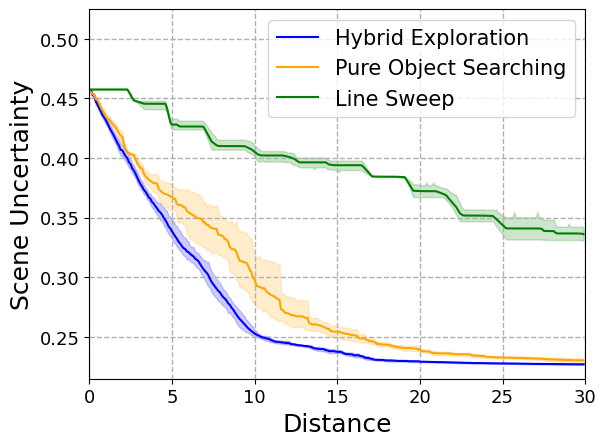}}
    \subfloat[Scene \textcircled{4}, $U^C$]{\includegraphics[width=0.48\linewidth]{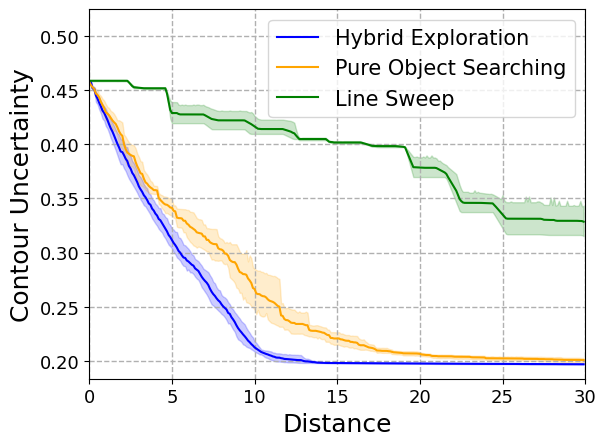}}
    \caption{Evaluation metric $U^S$ and $U^C$ versus the total travel distance in two different scenes. {\revcolor{red}The shadowed regions are the 95\% confidence intervals.}}
    \label{fig:metric_curve}
\end{figure}

The simulation experiment was conducted over 6 scenes. {\revcolor{red}In those experiments, the sensor's contact model is a point. The contact observations are noise-free i.e., exactly on the surface of the object. The starting point of exploration is randomized.} The results of the object contour extraction are presented in Fig.~\ref{fig:reconstruct}. Particularly, it shows the quantitative efficiency results of two scenes: 1) an environment with 5 small objects (Fig.~\ref{fig:reconstruct} \textcircled{1}), and 2) a scene with non-convex objects with inter-occlusion (Fig.~\ref{fig:reconstruct} \textcircled{4}). {\revcolor{red}These two scenes are chosen for the efficiency benchmark since they are identified to be the most representative scenes. This is because scene \textcircled{1} and \textcircled{4} contains all test cases concerned i.e., scattered small objects, narrow corridors, non-convex objects, occlusion. The efficiency benchmark results for other scenes can be found in Sec.~\ref{sec:sp}.} The efficiency metrics $U^S$ and $U^C$ as a function of the trajectory length are shown in Fig.~\ref{fig:metric_curve}. The mean value (curves), as well as 95\% confidence interval (shadow regions) calculated from 5 experiments per algorithm in each scene, are presented in Fig.~\ref{fig:metric_curve}.

First, it can be seen that when the hybrid policy (blue curve) was used, both $U^S$ and $U^C$ were reduced at a faster rate than the line sweep algorithm baseline \cite{kaboli2019tactile} (green curve). This superior performance can still be noticed even when the contour tracing was removed (orange curve), showing the efficiency advantages in object searching. Next, the contour tracing can improve the metric $U^{C}$, as reflected by the results that hybrid policy can outperform two other baseline algorithms in Fig.~\ref{fig:metric_curve} (b) and (d). In addition, it can also be observed that the confidence interval of the hybrid policy is smaller than other baselines in scene \textcircled{4}. The reason is that hybrid policy can maintain a consistent movement pattern in contact-rich scenes with a stable exploration rate. In comparison, the paths from two baseline policies were frequently interrupted by collisions, and have to be re-planned.

\begin{figure}[htb]
    \centering
    \subfloat[Scene \textcircled{1}]{\includegraphics[width=0.23\linewidth,trim=100 38 90 30,clip]{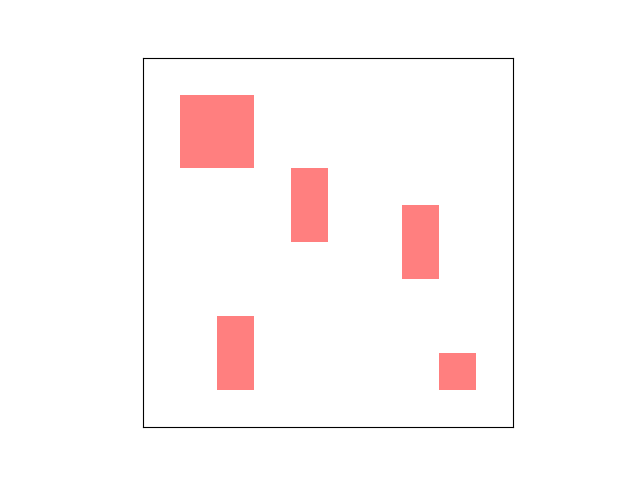}}
    \subfloat[(1)]{\includegraphics[width=0.23\linewidth,trim=40 40 40 40,clip]{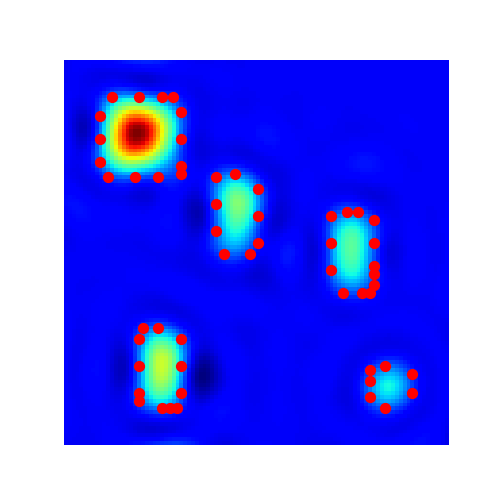}}
    \subfloat[(2)]{\includegraphics[width=0.23\linewidth,trim=40 40 40 40,clip]{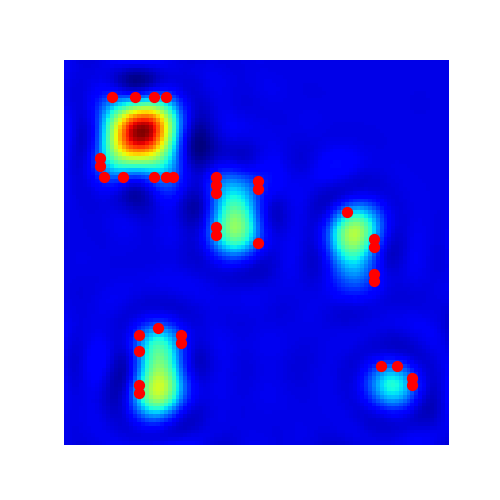}}
    \subfloat[(3)]{\includegraphics[width=0.23\linewidth,trim=40 40 40 40,clip]{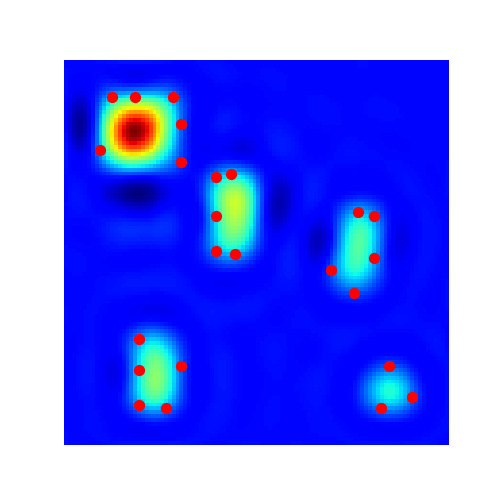}}

    \subfloat[Scene \textcircled{4}]{\includegraphics[width=0.23\linewidth,trim=100 38 90 30,clip]{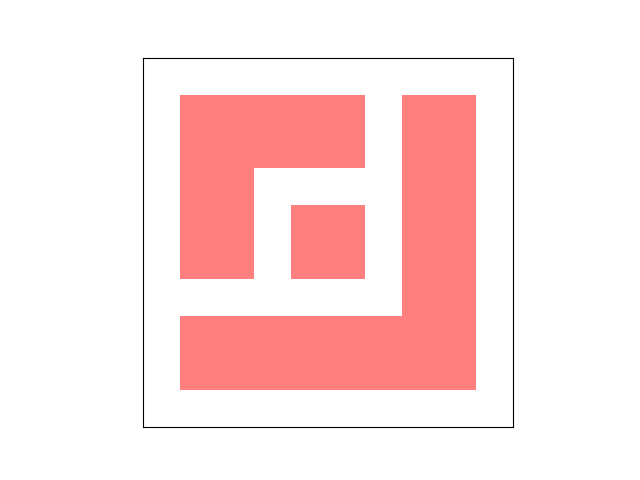}}
    \subfloat[(1)]{\includegraphics[width=0.23\linewidth,trim=40 40 40 40,clip]{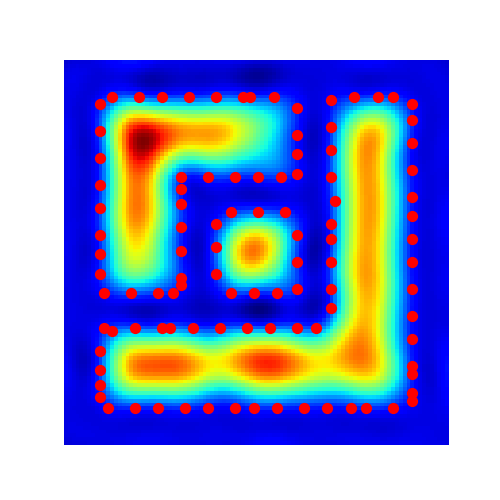}}
    \subfloat[(2)]{\includegraphics[width=0.23\linewidth,trim=40 40 40 40,clip]{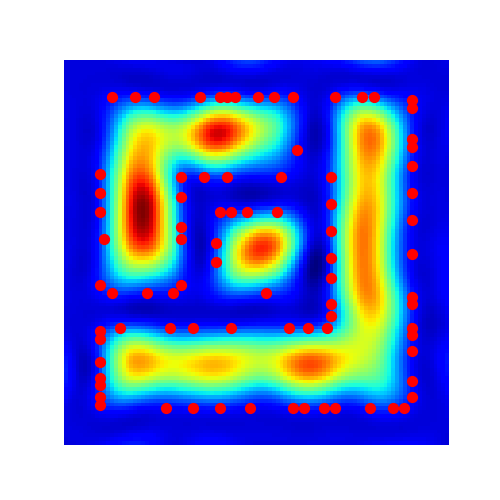}}
    \subfloat[(3)]{\includegraphics[width=0.23\linewidth,trim=40 40 40 40,clip]{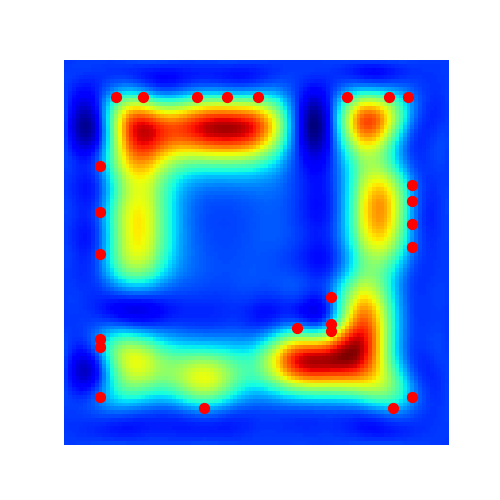}}
    
    \caption{Qualitative comparison of the exploration results. The left-most column shows the ground-truth layout of scene \textcircled{1} and \textcircled{4}. (1) Hybrid policy, (2) baseline 1: pure object searching policy, (3) baseline 2: line sweep policy \cite{kaboli2019tactile}. The color indicates the likelihood occupancy predicted by Gaussian Process. Dots are contact points.}
    \label{fig:qual_30}
\end{figure}

Fig.~\ref{fig:qual_30} shows plots of the contact points, and the predicted values of occupancy function $\hat{f}$ using the Gaussian Process at the end of this exploration session. First, the contact points from the hybrid policy are distributed evenly around the contour by virtue of the contour tracing policy. This effect cannot be seen on the two other baselines. Second, the hybrid and the pure object searching baseline policy enabled successful object localization even when inter-occlusion exists between objects, as shown in scene \textcircled{4} in Fig.~\ref{fig:qual_30}. This is by virtue of the dense coverage guarantee from Proposition 2. In contrast, the line sweep baseline algorithm  failed to localize the square in the middle of the scene \textcircled{4}, which is occluded by two L-shape objects.

\captionsetup[subfigure]{labelformat=empty}
\captionsetup[subfigure]{labelformat=parens,aboveskip=0pt}

\begin{figure}[tb]
    \centering 
    \subfloat[$d=1.87$]{\includegraphics[width=0.245\linewidth,trim=20 0 20 0,clip]{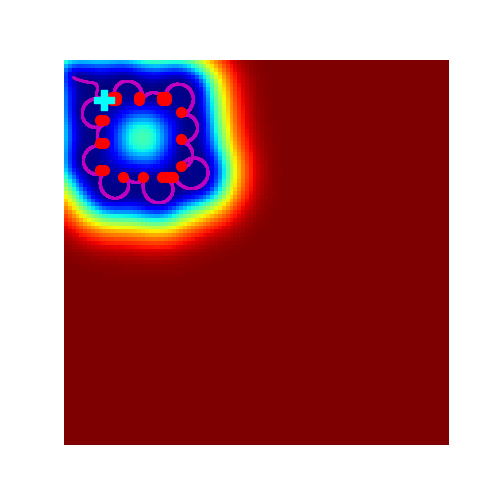}}
    \subfloat[$d=3.79$]{\includegraphics[width=0.245\linewidth,trim=20 0 20 0,clip]{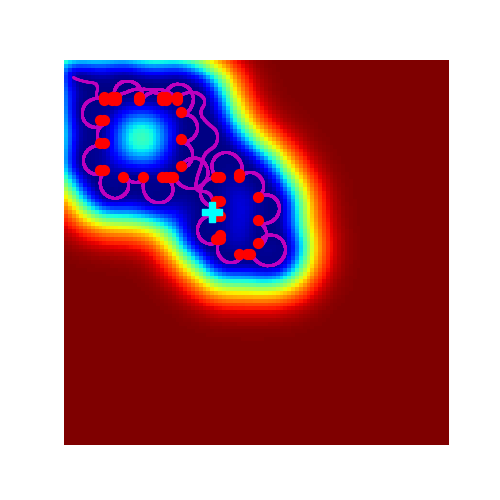}}
    \subfloat[$d=5.59$]{\includegraphics[width=0.245\linewidth,trim=20 0 20 0,clip]{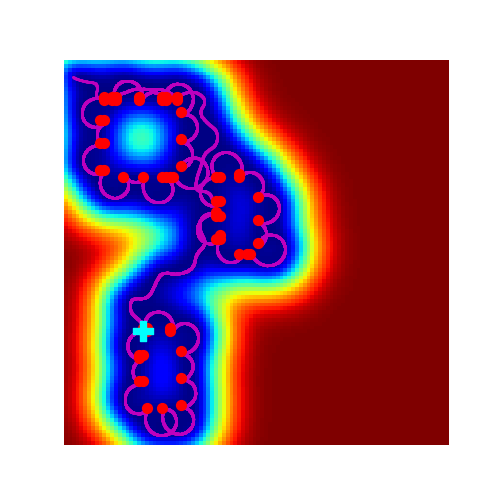}}
    \subfloat[$d=6.70$]{\includegraphics[width=0.245\linewidth,trim=20 0 20 0,clip]{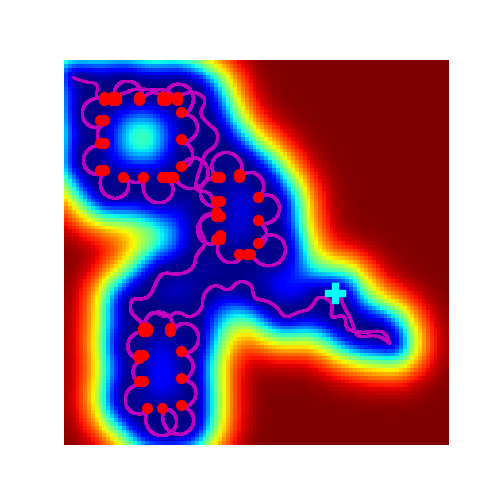}}
    
    \subfloat[$d=8.89$]{\includegraphics[width=0.245\linewidth,trim=27 0 27 0,clip]{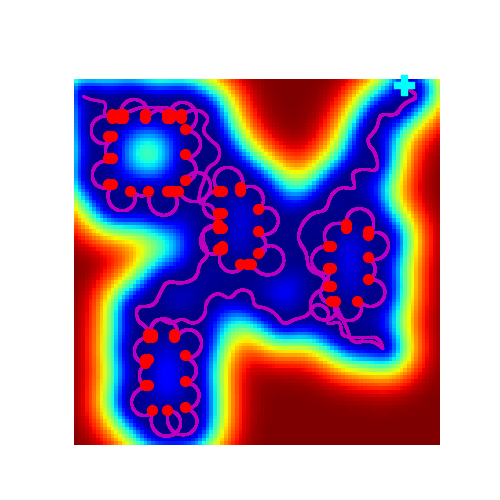}}
    \subfloat[$d=10.97$]{\includegraphics[width=0.245\linewidth,trim=20 0 20 0,clip]{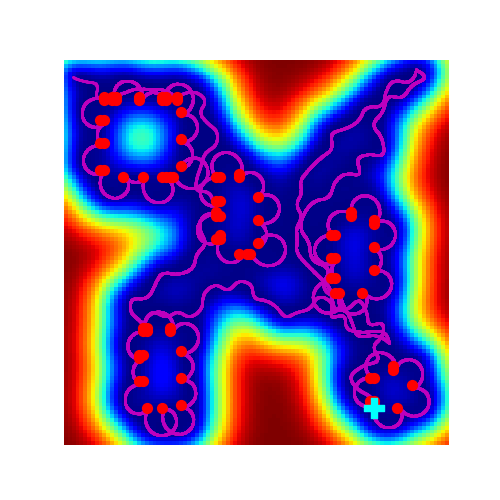}}
    \subfloat[$d=15.05$]{\includegraphics[width=0.245\linewidth,trim=24 0 24 0,clip]{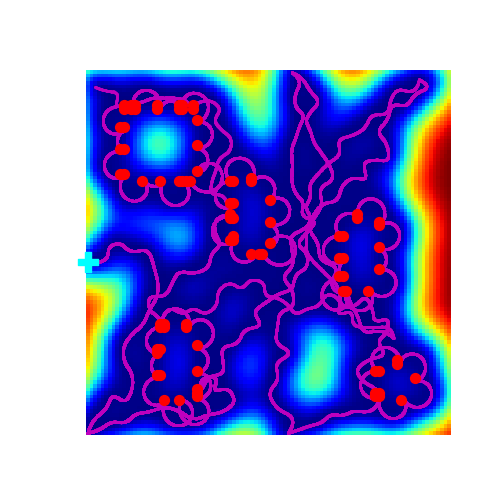}}
    \subfloat[$d=39.99$]{\includegraphics[width=0.245\linewidth,trim=24 0 24 0,clip]{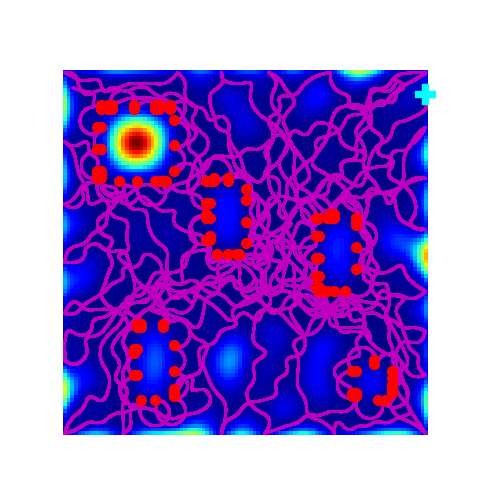}}
    \caption{Exploration procedure of scene \textcircled{1}. The cyan ``+'' is the position of the robot. The background color is the heatmap of standard deviation. Magenta trajectory indicates the sensor's traveling history. Red dots are contact points.}
    \label{fig:scene2explore}
    \subfloat[$d=1.19$]{\includegraphics[width=0.245\linewidth,trim=20 0 20 0,clip]{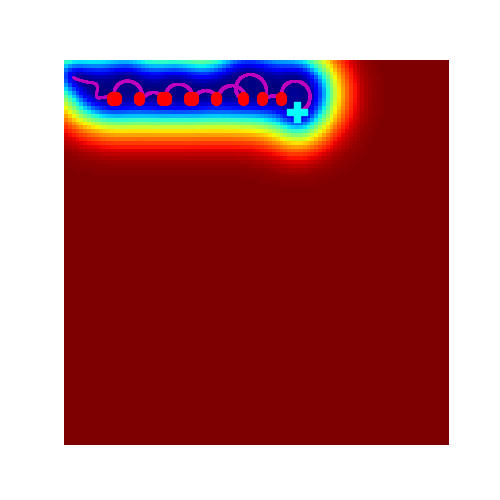}}
    \subfloat[$d=1.90$]{\includegraphics[width=0.245\linewidth,trim=20 0 20 0,clip]{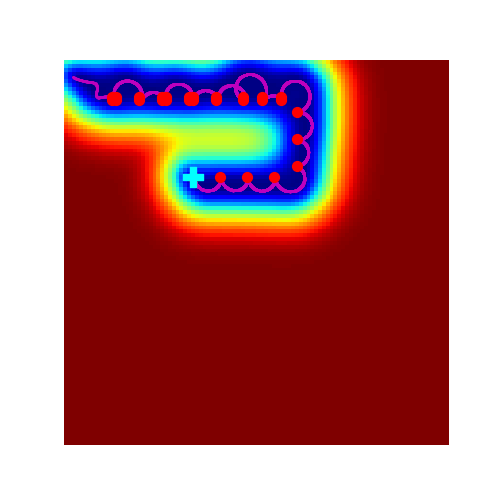}}
    \subfloat[$d=3.64$]{\includegraphics[width=0.245\linewidth,trim=20 0 20 0,clip]{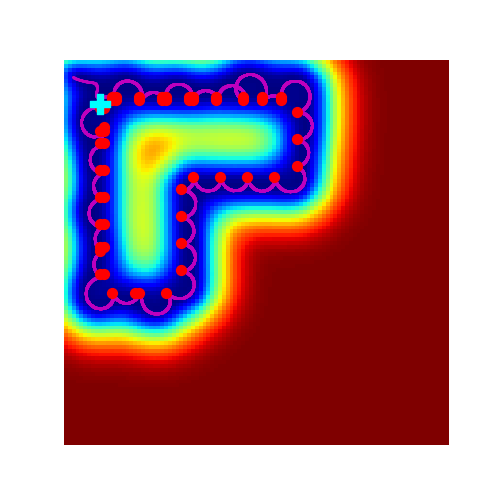}}
    \subfloat[$d=5.48$]{\includegraphics[width=0.245\linewidth,trim=20 0 20 0,clip]{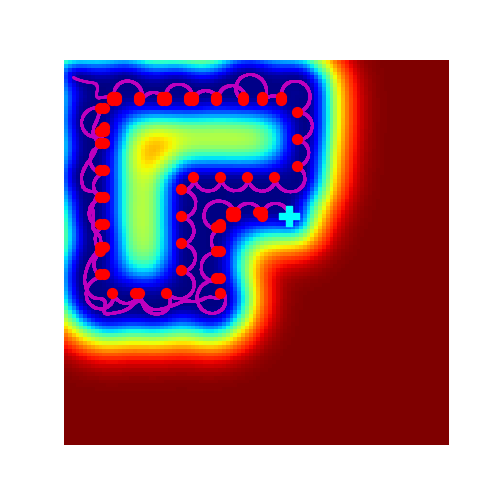}}
    
    \subfloat[$d=6.65$]{\includegraphics[width=0.245\linewidth,trim=20 0 20 0,clip]{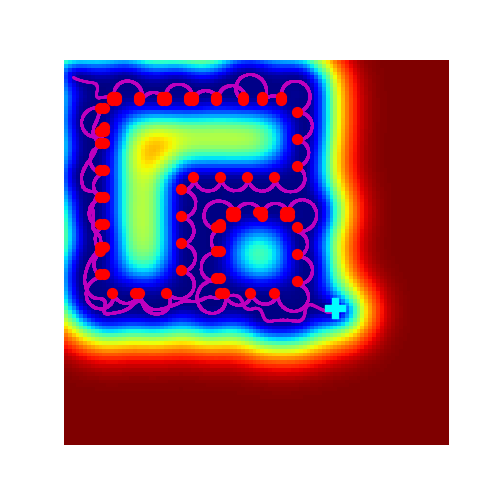}}
    \subfloat[$d=8.21$]{\includegraphics[width=0.245\linewidth,trim=20 0 20 0,clip]{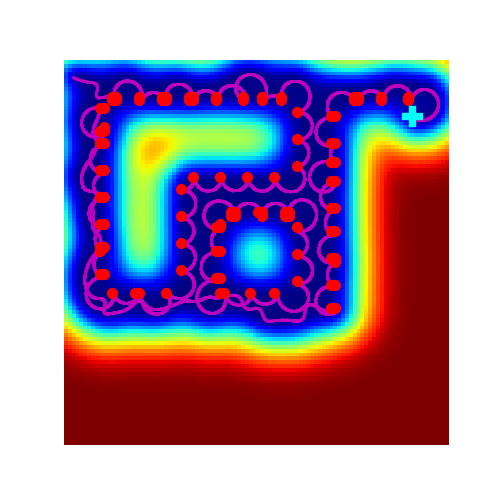}}
    \subfloat[$d=10.81$]{\includegraphics[width=0.245\linewidth,trim=20 0 20 0,clip]{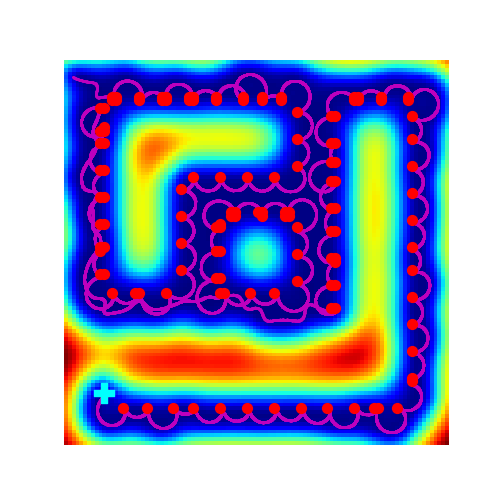}}
    \subfloat[$d=12.38$]{\includegraphics[width=0.245\linewidth,trim=20 0 20 0,clip]{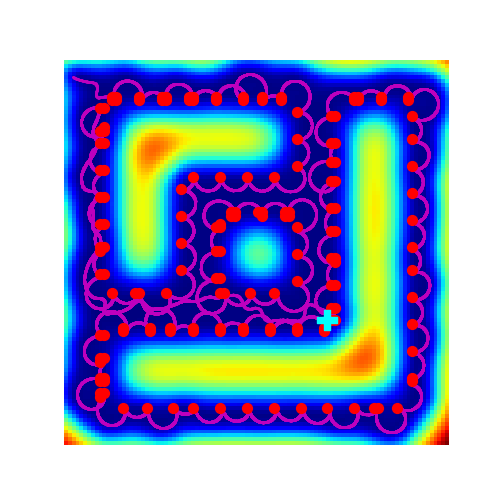}}
    \caption{Exploration procedure of scene \textcircled{4}. The color protocol is the same as Fig.~\ref{fig:scene2explore}. }
    \label{fig:scene3explore}
\end{figure}

Fig.~\ref{fig:scene2explore} and Fig.~\ref{fig:scene3explore} show the changes of $h(\mathbf{x})$ over travel distance, and the trajectory of the hybrid policy, respectively. There are two facts that can be observed from these figures. First, the region area covered by exploration paths increased with the travel distance. Note that at the end of this exploration session, the regions with high uncertainty only remained inside the contour, which cannot be visited by the sensor. Second, the hybrid policy avoided planning paths crossing through the discovered objects by explicitly using their contour polygons. This is opposed to the two baselines where collision paths are unavoidable if the contact observations do not fully cover the contour.

\captionsetup[subfigure]{labelformat=empty,aboveskip=-3pt}
\captionsetup[subfigure]{labelformat=parens,aboveskip=0pt}

\subsection{Contour Reconstruction} \label{localization_acc}

Object contour can be extracted by the method proposed in Sec.~\ref{met:ol}. In Fig.~\ref{fig:reconstruct}, the contour extraction results from 6 different scenes are demonstrated.  To be specific, we tested on primitive shapes in scenes \textcircled{1}, \textcircled{2}, and \textcircled{3}. In particular, the size of objects in scenes \textcircled{1} and \textcircled{3} are relatively small, and thus difficult to be found during object searching. Even though the challenge, all objects can be successfully localized. For small objects, contact points in scene \textcircled{3} may not be sufficiently dense to represent the ground truth shape accurately, because the oscillator radius $r_H$ used is relatively large compared to the object size (squares, with a side length of 0.09). This issue can be solved by choosing a smaller value for $r_H$ when detecting small objects, at a cost of longer traveling distance due to more contact events.  

Scenes \textcircled{4}\textcircled{5}\textcircled{6} are experiments with non-convex objects. For scene \textcircled{4}, the gap distances between objects are relatively small compared to the object scale. For this reason, it is intractable to segment different objects by clustering correctly. In comparison, our object extraction method (Sec.~\ref{met:ol}) can succeed in reconstructing all three objects. In scene \textcircled{5}, the hybrid policy succeeds in localizing all pentagrams. But similar to \textcircled{2}, improving the reconstruction of those sharp corners of the pentagrams requires smaller $r_H$. Last, the results from scene \textcircled{6} show that the Hopf oscillator can go in and out of small chambers in the ``S'' shape, showing the robustness of the contour tracing motion. 

\subsection{Experiment with Whiskers on the Robot} \label{sec:real1}

\begin{figure}[htb]
    \centering
    \subfloat[]{\includegraphics[width=0.29\linewidth, height=0.29\linewidth]{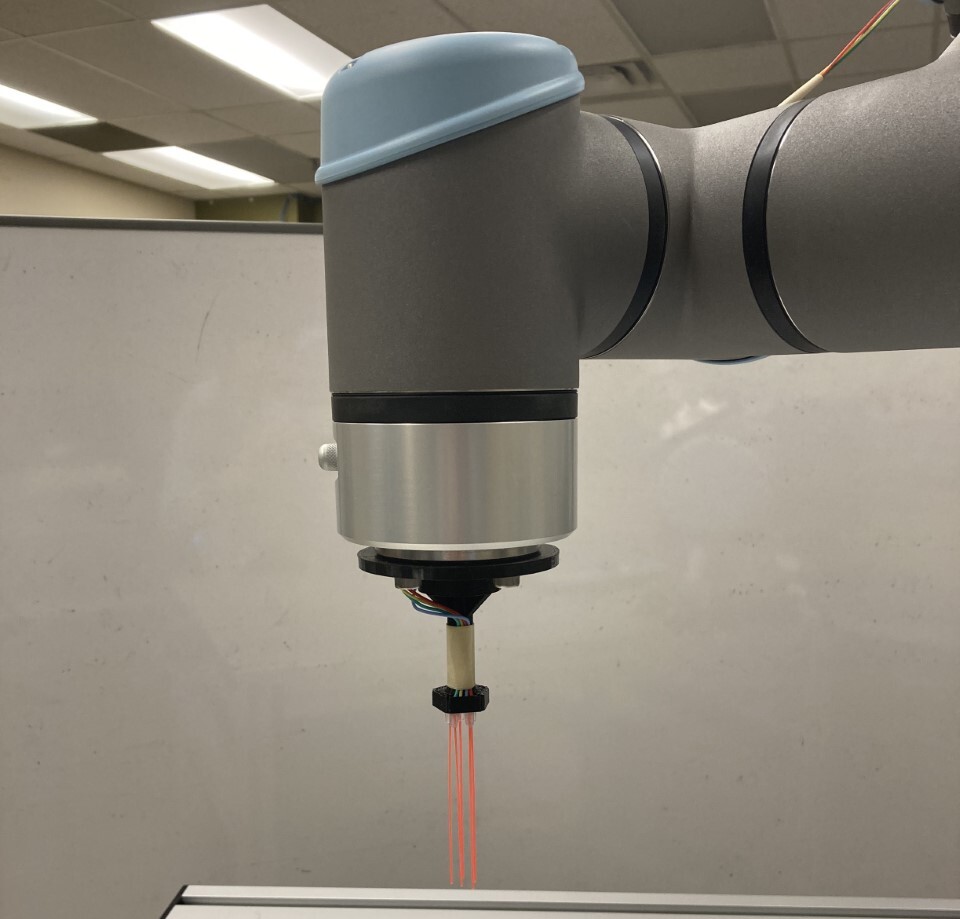}}\hspace{0.01\linewidth}
    \subfloat[]{\includegraphics[width=0.29\linewidth, height=0.29\linewidth]{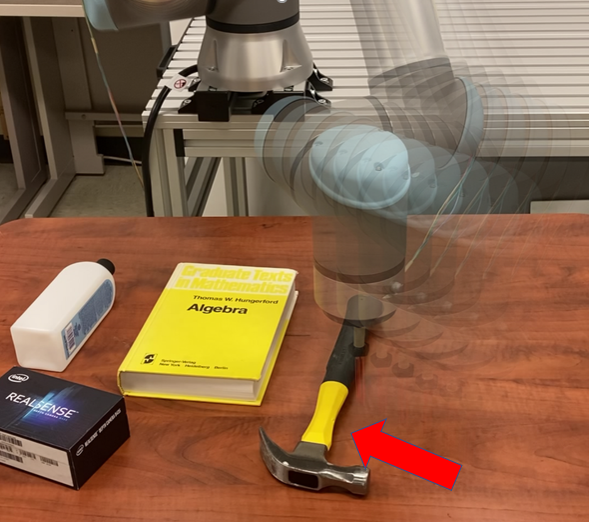}}\hspace{0.01\linewidth}
    \subfloat[]{\includegraphics[width=0.29\linewidth, height=0.29\linewidth]{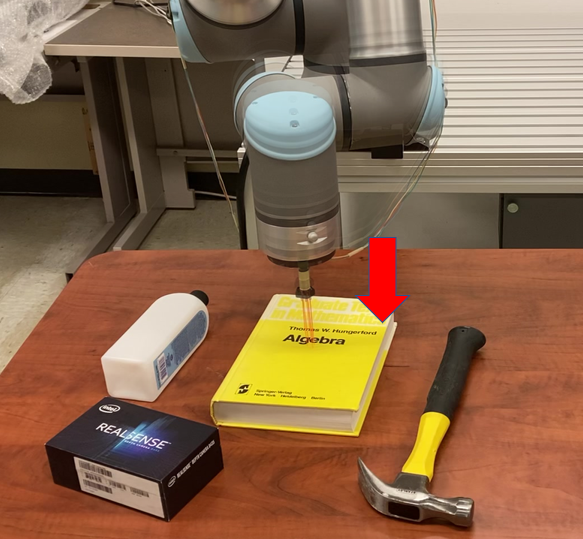}}

    \subfloat[]{\includegraphics[width=\linewidth]{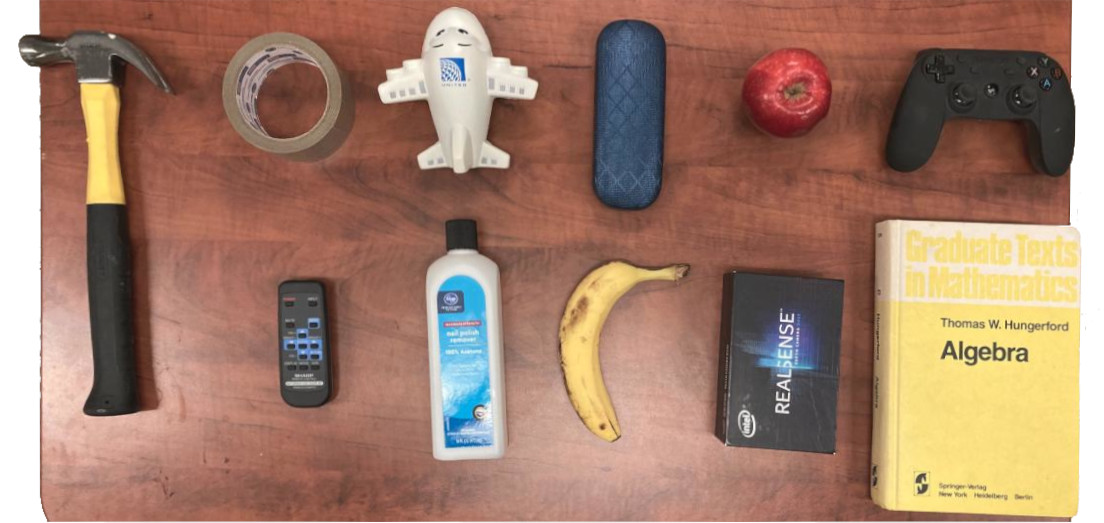}}
    
    \caption{(a) Whisker sensor mounted on a UR16e robot, (b) object searching on the desktop, (c) feature sampling by probing from above of the objects, (d) objects used in the experiments. }
    \label{fig:UR_robot_mount}
\end{figure}

The proposed approaches were implemented on a UR16e robot. The tactile feedback was obtained by using the whisker sensor developed, which was mounted onto the robot's end-effector using a 3D-printed flange, as shown in Fig.~\ref{fig:UR_robot_mount} (a). During object searching and contour tracing, the robot moved in a horizontal plane ({\revcolor{red}$z=0.01\;m$, where the desk surface is defined as $z=0\;m$}), as shown in Fig.~\ref{fig:UR_robot_mount} (b). Once the object is being localized, it is possible to sample points directly in volumetric space $\mathbb{R}^3$ in order to enhance the object feature discriminability. This was done by approaching from top of the object until contact events, as shown in Fig.~\ref{fig:UR_robot_mount} (c). {\revcolor{red} Note that, if the center lies outside of the object contour, the measured $z$ is $0\;m$.} The experiments included 11 real objects, which are shown in Fig.~\ref{fig:UR_robot_mount} (d). The real objects include: a tape, a rubber airplane, an eyeglasses case, an apple, a game controller, a TV remote control, a bottle, a banana, a Realsense camera package box, and a book. The objects chosen have different masses, dimensions, affordances, materials and textures, friction coefficients, and convexities. {\revcolor{red}Since the area of the task space was limited by the reachability of the robot arm, 3-6 objects were selected for each experiment. }

\begin{figure}[htb]
    \centering
    \includegraphics[width=0.99\linewidth]{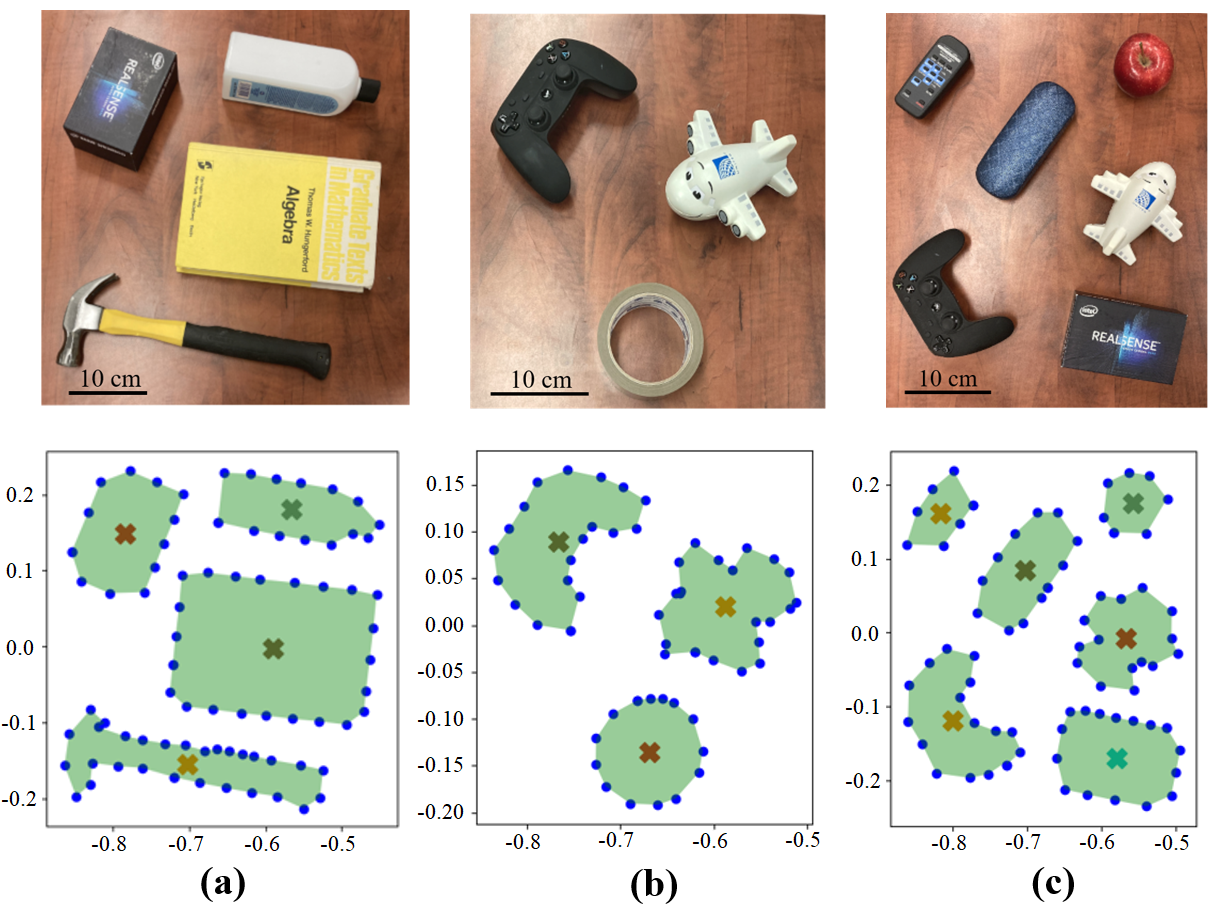}

    \vspace{-2mm}
    \caption{Results of active tactile exploration in real experiments. The object contours in 3 different real scenes (the 1st row) were reconstructed by only tactile feedback (the 2nd row). Marker \ding{54} is the estimated centroid of each object.}
    \label{fig:real_object_distribution}
\end{figure}

\begin{figure}[htb]
    \centering
    \subfloat[$d=1.20$]{\includegraphics[width=0.24\linewidth,trim=40mm 10mm 40mm 10mm,clip=true]{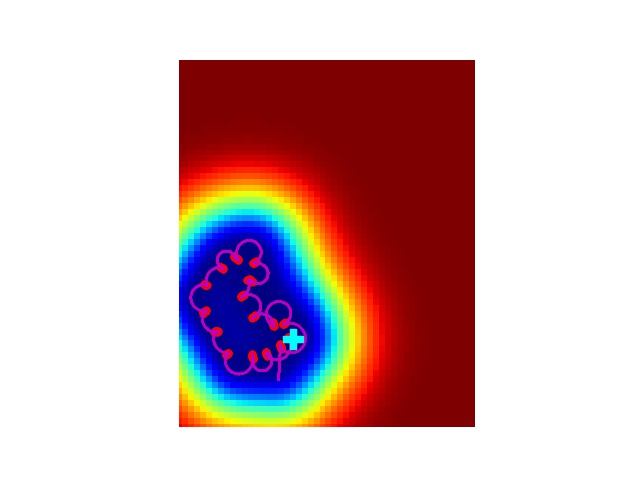}}
    \subfloat[$d=2.61$]{\includegraphics[width=0.24\linewidth,trim=40mm 10mm 40mm 10mm,clip=true]{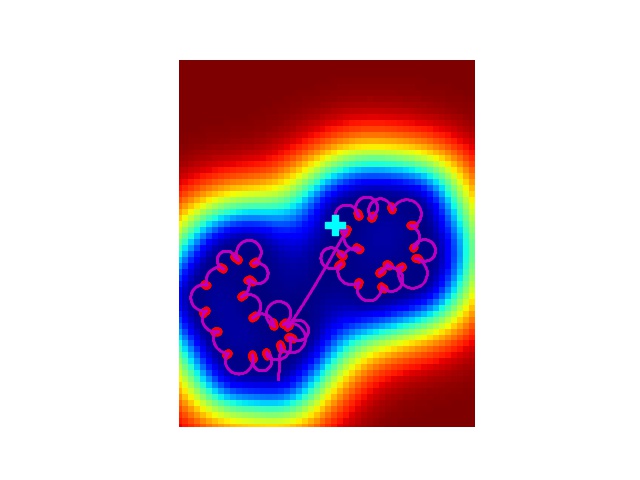}}
    \subfloat[$d=3.46$]{\includegraphics[width=0.24\linewidth,trim=40mm 10mm 40mm 10mm,clip=true]{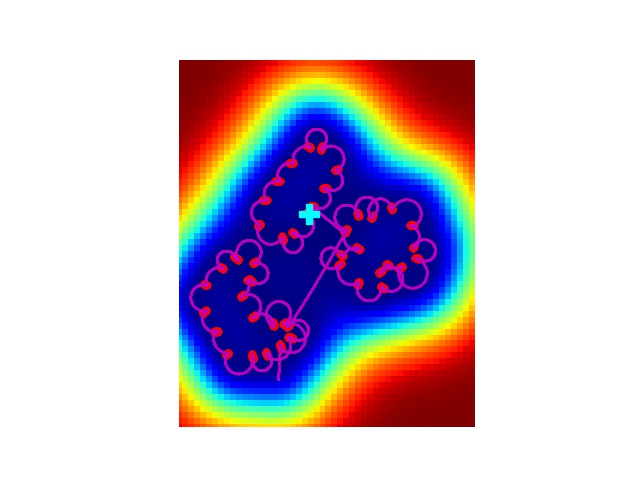}}
    \subfloat[$d=4.14$]{\includegraphics[width=0.24\linewidth,trim=40mm 10mm 40mm 10mm,clip=true]{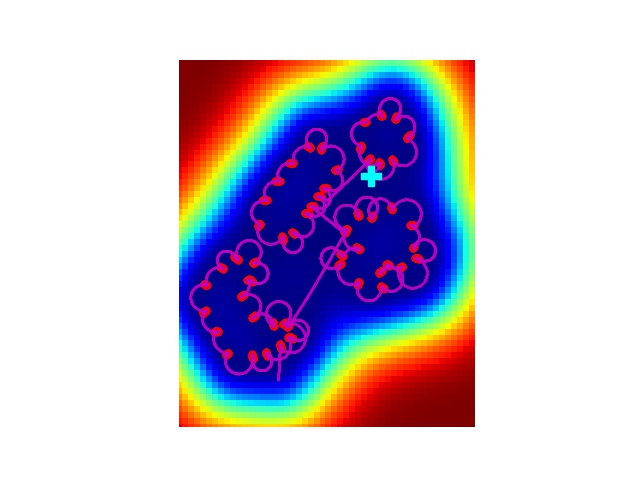}}
    
    \subfloat[$d=6.43$]{\includegraphics[width=0.24\linewidth,trim=40mm 10mm 40mm 10mm,clip=true]{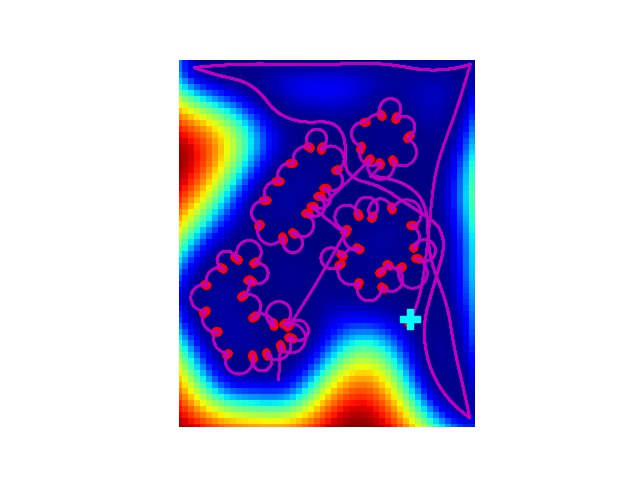}}
    \subfloat[$d=7.83$]{\includegraphics[width=0.24\linewidth,trim=40mm 10mm 40mm 10mm,clip=true]{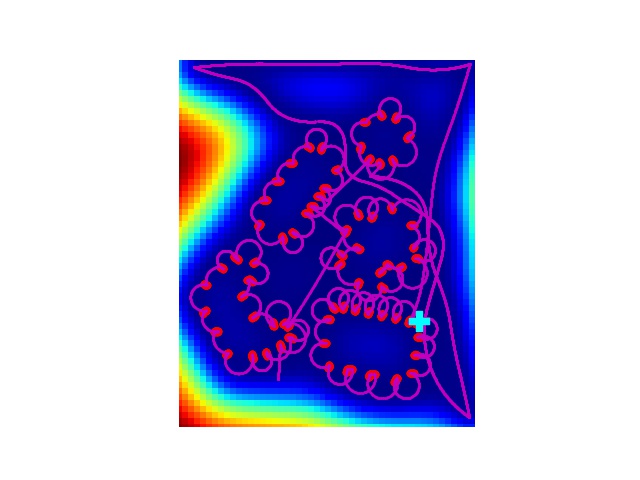}}
    \subfloat[$d=8.92$]{\includegraphics[width=0.24\linewidth,trim=40mm 10mm 40mm 10mm,clip=true]{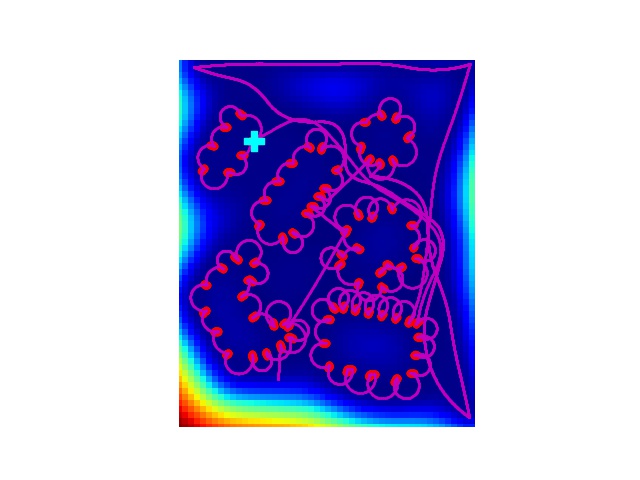}}
    \subfloat[$d=10.69$]{\includegraphics[width=0.24\linewidth,trim=40mm 10mm 40mm 10mm,clip=true]{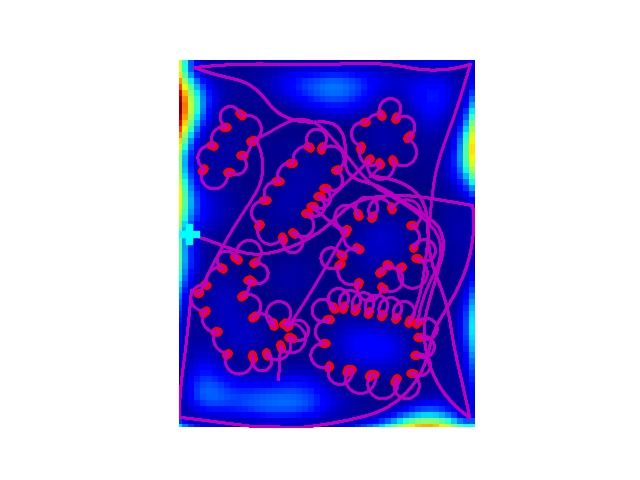}}
    \caption{The exploration procedures of the real scene in  Fig.~\ref{fig:real_object_distribution} (c). }
    \label{fig:real_traj}
\end{figure}

It was found that 60-120 contact events in each real experiment were required to characterize the objects.  To accomplish this, the contact detection procedure had to be robust, while at the same time non-disruptive (otherwise may reposition the object, which may deform the observed contour and lead to misrecognition). This indicates that our system is sufficiently robust to handle all 11 objects regardless of their shapes, configurations and positions. Fig.~\ref{fig:real_object_distribution} shows the appearance of real objects in the first row, together with the contact samples and extracted polygon contours in the second row. It can be seen from the figure that there exists accurate shape correspondence between the real object layouts and their descriptor representations. In addition, there were no object repositioning events thanks to the compliance provided by the whiskers. We refer the reader to the link below Fig.~\ref{fig:teaser} for the video recording of the exploration process. The corresponding image snapshot is shown in Fig.~\ref{fig:real_traj} as well.

\begin{figure*}[htb]
    \centering
    \subfloat[bottle]{\includegraphics[height=0.125\linewidth]{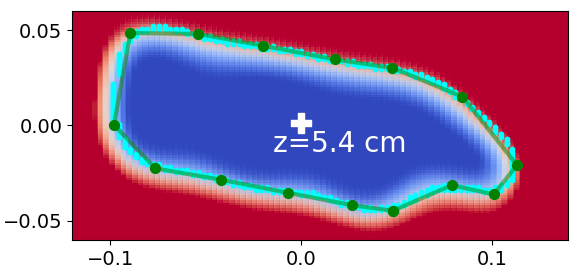}}
    \subfloat[hammer]{\includegraphics[height=0.125\linewidth]{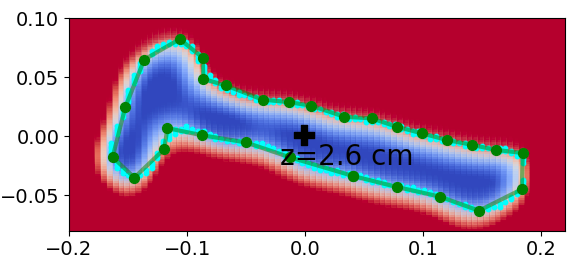}}
    \subfloat[rubber airplane]{\includegraphics[height=0.125\linewidth]{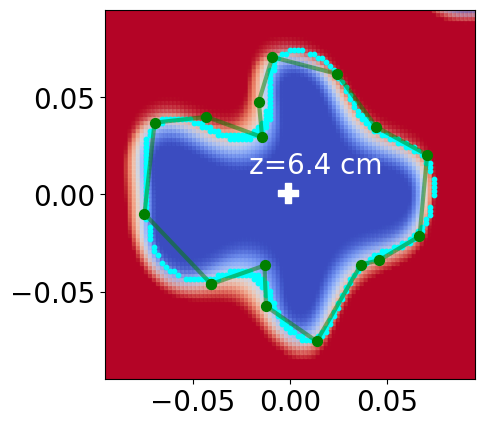}}
    \subfloat[banana]{\includegraphics[height=0.125\linewidth]{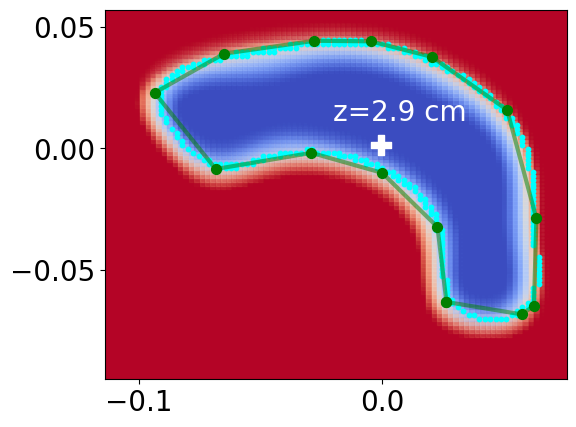}}
    \subfloat[apple]{\includegraphics[height=0.125\linewidth]{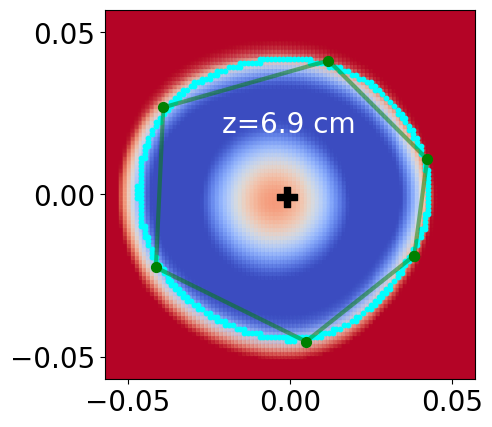}}
    
    \subfloat[eyeglasses box]{\includegraphics[height=0.125\linewidth]{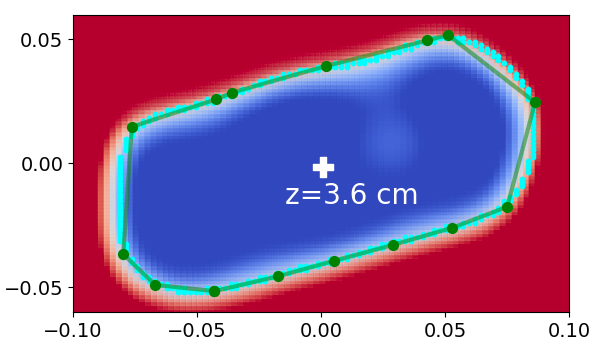}}
    \subfloat[tape]{\includegraphics[height=0.12\linewidth]{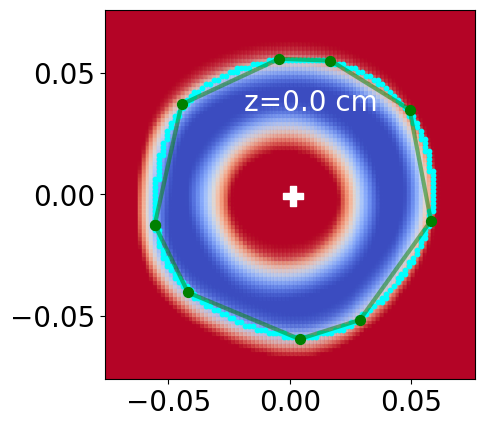}}
    \subfloat[Realsense package]{\includegraphics[height=0.12\linewidth]{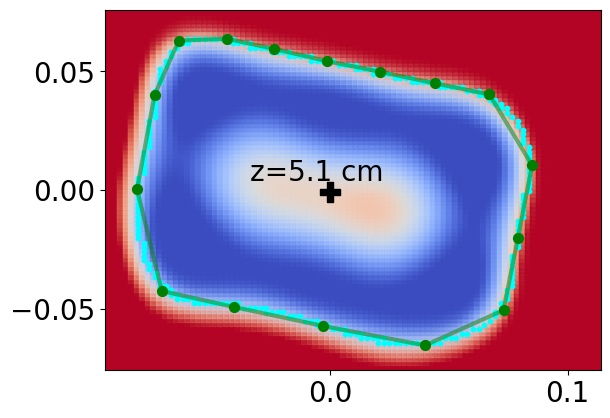}}
    \subfloat[TV control]{\includegraphics[height=0.12\linewidth]{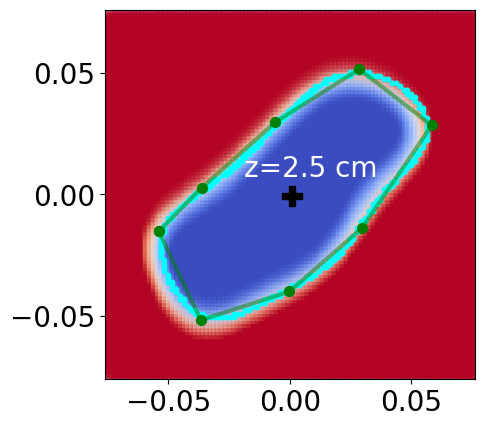}}
    \subfloat[book]{\includegraphics[height=0.12\linewidth]{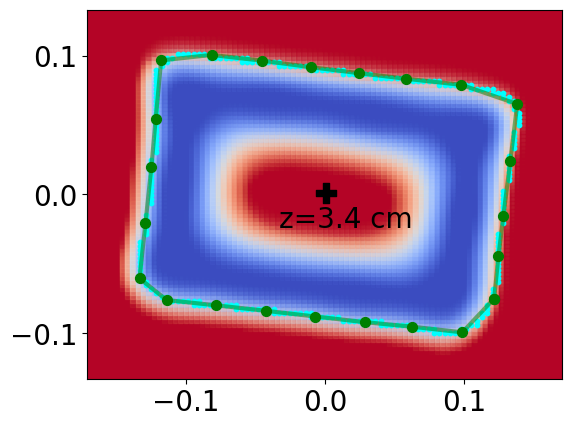}}
    \subfloat[game controller]{\includegraphics[height=0.12\linewidth]{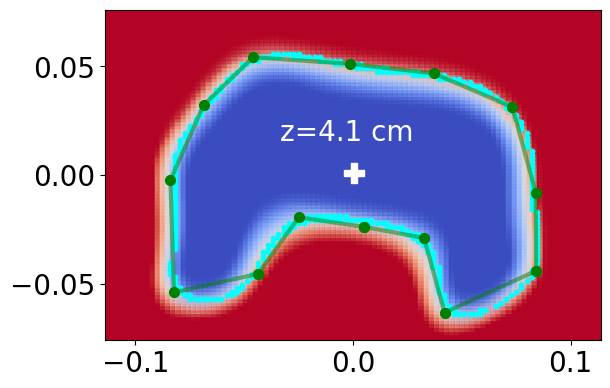}}
    \caption{{\revcolor{red}Examples of reconstructed contour polygons from 11 real objects. Green dots are contact points. \ding{58} corresponds to the centroid information with the measured height labeled. Cyan contour is the smoothed result from Gaussian Process implicit surface. Background color corresponds to the truncated signed distant field from the prediction.}}
    \label{fig:real_contours}
\end{figure*}

{\revcolor{red}
By leveraging Gaussian Process implicit surface \cite{williams2006gaussian}, the contour polygon is smoothed. This is shown in Fig.~\ref{fig:real_contours}, with the smoothed contour shape highlighted. In comparison, the raw contour polygon corresponds to the green curve. Note that there may be multiple 0-isosurfaces when there exist large uncertain regions without contact observations within them. Here only the isosurface with the smallest average distance to the raw contour polygon is displayed. 
}

\captionsetup[subfigure]{labelformat=empty,aboveskip=-1pt}
\captionsetup[subfigure]{labelformat=parens,aboveskip=0pt}

\section{Object Classification Results} \label{sec:res_or}

\subsection{Real-world object classification} \label{sec:realworld}

In this experiment, object contours from real objects were classified using the deep neural network introduced in Sec.~\ref{sec:classifier}. The data was collected using the same method as Sec.~\ref{sec:real1}, {\revcolor{red}in which objects were arranged as scenes first and were then characterized by the TOS-IPP algorithm. As a result, 4-10 observations were collected for each object, leading to 79 observations in total. The number of observations was determined by the available space for objects in the scene. For a constrained area in space, such an area can be filled with several small objects and fewer larger objects.} Fig.~\ref{fig:real_contours} shows one example observation per object. 
To use batch training, the number of points in $P_{ct}$ needs to be the same among different object observations. Thus, $P_{ct}$ was interpolated to 64 points at an equal distance interval based on the polygon obtained in the previous step. 

The ability to learn from scarce observations (data efficiency) is important, since the acquisition of a large dataset of observations from real scenes is expensive.
We first quantified the data efficiency by training the proposed network with a reduced amount of data. Only approximately 20\% instances in the dataset were used for training, and the rest were used for validation (Setting A). The results were then compared to setting B, in which around 80\% of the observations were used for training and the remaining 20\% for validation. To reduce the performance variance, the network was evaluated by 5-fold cross-validation. The final metric is the average accuracy of all folds. Last, an ablation study was conducted to quantify the contribution from $P_{fs}$. For this, the accuracies of the following two settings were compared: 1) training and validating the network with only $P_{ct}$, and 2) training and validating the network with $P_{ct}$ and $P_{fs}$ altogether. 

\begin{table}[htb]
\caption{Validation accuracy (in \%) using the dataset created from real objects.}
\vspace{-3pt}
\label{tab:realobjeval}
\centering
\setlength{\tabcolsep}{1.5mm}{
\begin{tabular}{@{}llllllll@{}}\hline\hline
\multicolumn{2}{l}{Experiment Setting}      & Fold 1 & Fold 2 & Fold 3 & Fold 4 & Fold 5 & Avg\\ \hline
\multirow{2}{*}{\;\;\textbf{A}} & $P_{ct} \cup P_{fs}$  & \textbf{74.6}   & \textbf{95.2}   & \textbf{92.2}  & \textbf{88.9} & \textbf{62.7} & \textbf{82.7}\\ 
                    & $P_{ct}$ only &  71.2  & 92.1   & 81.2  & \textbf{88.9} & 61.2 & 78.9\\ \hline
\multirow{2}{*}{\;\;\textbf{B}} & $P_{ct} \cup P_{fs}$   &  \textbf{100.0}   & \textbf{100.0} & \textbf{100.0} & \textbf{100.0} & \textbf{91.7}  & \textbf{98.3}\\ 
                    & $P_{ct}$ only &  \textbf{100.0}   & 93.8 & \textbf{100.0} & \textbf{100.0}  & \textbf{91.7} & 97.1  \\ \hline\hline
\end{tabular}}
\end{table}

The results of these trials can be found in Table.~\ref{tab:realobjeval}. In this table, each entry is a classification accuracy obtained from an individual validation session. The network was trained using a learning rate at $1\times10^{-4}$ for 500 episodes with data augmentation. The data augmentation was accomplished by rotating the contour around its centroid point by a random angle (sampled from a uniform distribution between $[0, 2\pi)$). It can be observed that the CT-Net is effective, as when using only 20\% observation instances for training (setting A, with one $\mathbb{R}^3$ point in $P_{fs}$), the averaged validation accuracy was 82.7\%. This accuracy was increased to 98.3\% when the training split was increased to 80\% instances. We conclude that the proposed network is data-efficient, because each fold of training data in setting A only includes 12-20 observations. Such amount of data is considerably small regarding the number of categories is 11. Besides, it can be observed that the validation accuracy in the $P_{ct} \cup P_{fs}$ setting achieves equivalent or better performance than using only $P_{ct}$ in all cases. This proves that using the $\mathbb{R}^3$ point set $P_{fs}$ can improve the object discriminability. %

\subsection{Scalability and Ablation Studies} \label{sec:virtual_obj_recongition}

An experiment was conducted to show that the proposed classification network is capable to tackle classification problems with a larger dataset of observation instances and categories. For this, we utilize the 3DNet dataset \cite{wohlkinger20123dnet}, which is composed of aligned mesh models from 222 categories. Because the categorical sample distribution in 3DNet is highly imbalanced, only categories with more than 15 objects were used. That is, a total of 2098 objects that belong to 68 categories were adopted. 70\% of those objects were split for training and the remaining 30\% for validation.  

\begin{figure}[htb]
    \centering
    \subfloat[]{\includegraphics[width=0.3\linewidth]{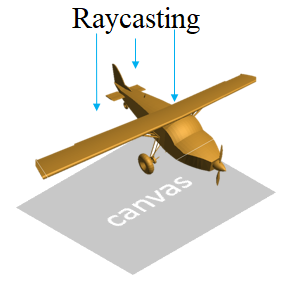}}
    \subfloat[]{\includegraphics[width=0.3\linewidth]{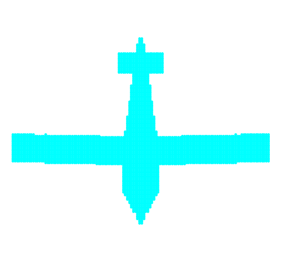}}
    \subfloat[]{\includegraphics[width=0.3\linewidth]{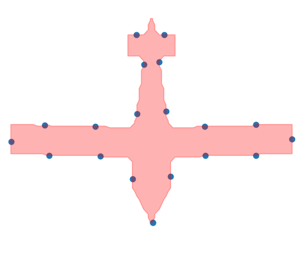}}
    \caption{(a) Demonstration of raycasting, (b) planar projection of the mesh model, (c) concave hull polygon calculated by $\alpha$-shape, and sampled points from this polygon.}
    \label{fig:procedure_examples_raycasting}
\end{figure}

Due to the availability of mesh models, the points in the object contour set $P_{ct}$ were directly obtained by raycasting without the need of measurements by physical interactions. This was done by three steps, as shown in Fig.~\ref{fig:procedure_examples_raycasting}. First, the surface vertices were projected onto a planar canvas at $z=0$. Second, the contour estimation $\hat{C}_i$ was generated by calculating the concave hull polygon through $\alpha$-shape algorithm \cite{edelsbrunner1983shape}. Last, a given number of points were sampled from the polygon to create the contour point set $P_{ct}$.  For $P_{fs}$, a given number of $\mathbb{R}^2$ points $(x^i_{fs}, y^i_{fs})$ were generated using the same method described in Sec.~\ref{sec:recog}. The $z^i_{fs}$ was then obtained at each point by raycasting from $z=+\infty$ in a line perpendicular to $z=0$, which is defined as the first intersection point between the ray and the mesh model. 

The network was trained for 200 episodes using the Adam optimizer. The learning rate was $2 \times 10^{-2}$ at the beginning, and was then decayed to half the value for every 30 episodes. The classification accuracies are shown in Table.~\ref{tab:3dneteval}. Overall, the highest classification accuracy achieved is 61.1\%. This proves the algorithm's discriminability considering that there exist 68 categories in total. This is also reflected by the comparison with PointNet \cite{qi2017pointnet}, which is a deep neural network designed for recognizing volumetric points only. For a PointNet model that was trained and tested with 64 volumetric points in $P_{fs}$ (Experiment \textcircled{7}), the accuracy achieved is 37.5\%, which is lower than the performances achieved by CT-Net.

An ablation study was conducted to investigate how the accuracy of CT-Net is affected by the number of contour points in  $P_{ct}$ and the number of volumetric points in $P_{fs}$. For $P_{ct}$, it was found that the accuracy was affected obviously when $P_{ct}$ had less than 25 points. Conversely, the influences decreased for the points added afterward. For $P_{fs}$, the accuracy increased significantly for the first observation (44.4\% $\longrightarrow$57.5\% with 50 points in $P_{ct}$), but the accuracy was less affected by adding more observations. For instance, it only gets 0.1\% accuracy boost when adding 63 more points afterward (\textcircled{2} and \textcircled{3}). To reduce the sampling cost, only one point was sampled for $P_{fs}$ in the real experiment (Sec.~\ref{sec:realworld}).

\begin{table}[htb]
\centering
\vspace{-3pt}
\caption{Classification accuracies (in \%) on the 3DNet dataset under different number of points in $P_{ct}$ and $P_{fs}$.}
\vspace{-3pt}
\label{tab:3dneteval}
\begin{tabular}{ccccc}
\hline\hline
Index & Architecture & No. pts in $P_{ct}$ & No. pts in $P_{fs}$                   &  Accuracy (\%) \\\hline 
\textcircled{1} & CT-Net & 256 & 64 & 61.1\\
\textcircled{2} & CT-Net & 50 & 64 & 57.6\\
\textcircled{3} & CT-Net & 50& 1& 57.5\\
\textcircled{4} & CT-Net & 25& 1& 52.2\\
\textcircled{5} & CT-Net & 50& 0& 44.4\\
\textcircled{6} & CT-Net & 25 & 0 & 39.1\\\hline
\textcircled{7} & PointNet \cite{qi2017pointnet} & - & 64 &37.5\\
\textcircled{8} & PointNet \cite{qi2017pointnet} & - & 32 &35.8\\ %
\hline\hline
\end{tabular}
\vspace{-5pt}
\end{table}

\begin{figure}
    \centering
    \includegraphics[width=0.98\linewidth]{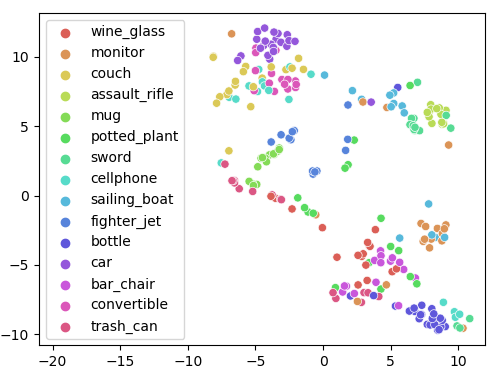}
    \caption{tSNE analysis on the 15 most frequently appeared categories in the validation split of the 3DNet dataset.}
    \label{fig:tSNE}
\end{figure}

The tSNE analysis was used to visualize discriminability between categories, as shown in Fig.~\ref{fig:tSNE}. The visualization was created using the validation split of 3DNet dataset, with the same neural network as setting \textcircled{3} in Table.~\ref{tab:3dneteval}. The data distribution indicates that samples within the same category tend to be spatially clustered together, while different categories tend to be spatially separated from each other.

\section{Discussion}
\begin{figure}[t]
    \centering
    \subfloat[]{\vspace{4pt}\includegraphics[height=0.42\linewidth, trim = 50 0 50 0]{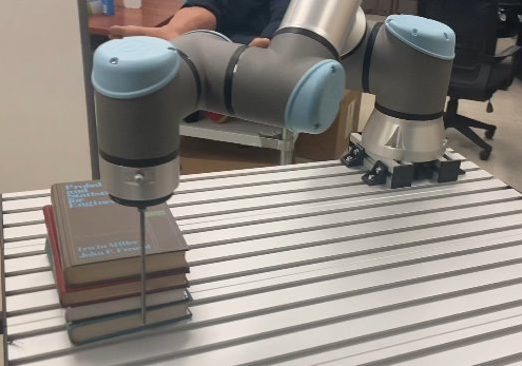}}
    \subfloat[]{\includegraphics[height=0.45\linewidth, trim = 50 0 50 0]{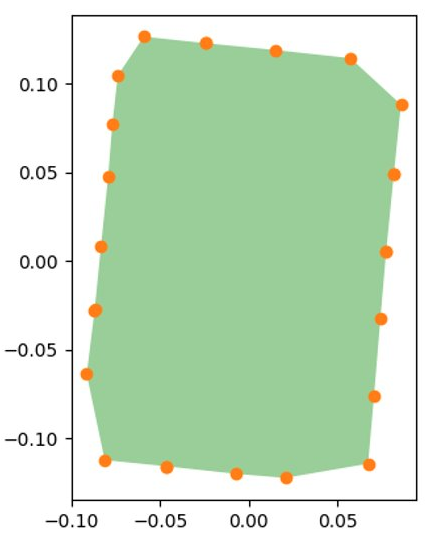}}
    \caption{(a) shape exploration by F/T sensor embedded in the UR16e robot's wrist, (b) contour shape obtained from the exploration. }
    \label{fig:book_exploration}
\end{figure}

The main objective of this work is to search and recognize objects by using the tactile modality alone, which allows gathering information from a scene when a vision modality is not available. Currently, our hardware system is based on a UR16e robot with a whisker sensor manufactured at a cost of around {\revcolor{red}\$15} each. Without loss of generality, the proposed approach can be applied to other robotic systems with at least a binary tactile sensor. {\revcolor{red}The algorithm was also tested based on the embedded 6-Axis Force/Torque (F/T) sensor of the UR16e robot. The scene exploration task was conducted with satisfactory results for heavy objects, such as a stack of books demonstrated in Fig.~\ref{fig:book_exploration}. However, due to the limited sensor's precision (5.5 N according to UR16e robot's datasheet), lightweight objects were frequently pushed away during the exploration (tested on the game controller, Realsense package box demonstrated in Fig.~\ref{fig:UR_robot_mount} (d)), leading to positional errors when estimating the object contour. This highlights the necessity of leveraging whiskers to achieve non-intrusive shape exploration.}

\subsection{Broader Applications}
{\revcolor{red}This work potentially has broad impacts on other spatial exploration applications as well. For instance, the same method can be used to approach the problem of finding hazardous objects concealed or buried, and recognizing their categories by the shape of contours without the need of extracting them out \cite{patel2020digger}.

While our sensor has not been integrated into manipulation tasks so far, the proposed exploration method could assist in manipulation tasks when vision is not available or effective. For instance, a bimanual robot with embedded whiskers on the dominant hand would use the non-dominant hand for blind grasping. This allows to search feasible grasps based on contour shape obtained during exploration with the dominant hand. Further, this would allow the robot to retrieve objects underwater, which is considered to be a challenging task when using the visual modality alone.} We believe that our techniques can solve these challenging problems in an efficient manner.

\subsection{Limitations}
The main limitation is the inability to operate in heavily cluttered scenes, where objects are in close contact with each other. {\revcolor{red}In those cases, the contour tracing would end up with a ``merged'' set of objects, which may lead to misrecognition. Recognizing a part of points from a merged 2D contour corresponds to the point cloud segmentation problem (i.e., attributing label to each point). A possible solution would be to leverage deep point cloud segmentation networks to propose candidate objects from the points, such as using architectures previously proposed \cite{qi2017pointnet, xiao2021triangle}. Another solution would be to use a planar pushing technique for separating the merged objects, such as in \cite{suresh2021tactile, sodhi2021learning}. On the flip side, pushing objects would require  to apply a significant force to the target objects, which is considered on itself an intrusive action. }

{\revcolor{red}
The spatial contact points for shape characterization are currently obtained by probing from the above of the object. However, those points  may only cover partial regions of an object when other regions are not reachable due to occlusion. A new exploration algorithm for a comprehensive surface
characterization is the subject of our ongoing work. The tangible region is enlarged by changing the whisker's pose. As a result, the obtained points could be used for object surface reconstruction. 
}

\section{Conclusion}

In this paper we studied the problem of how to utilize the tactile modality to explore, characterize, and understand the environment as well as the objects within it. Unlike many commercially used tactile sensors that may require a relatively large force or pressure to obtain tactile observations, our compliant whisker-based tactile sensor has high sensitivity and long sensing range. This allowed acquiring contact samples with minimal intrusiveness. The tactile intelligence is then introduced by designing tactile exploration policies, such as the hybrid exploration pattern proposed. This allows to actively search objects by planning informative paths, and reactively trace the object contour by making physical contacts.

The feasibility of the proposed methods was evaluated not only in simulation, but also on a real robot with the developed whisker sensor. The object contour, as well as the volumetric contact points were successfully obtained with real versatile daily objects. A deep neural network was used to classify these observations, which proves that the contour shape is informative enough about the object category. In addition, experiments were conducted to show that this classification approach is generalizable to large datasets. 

In the future, we plan to extend this framework to assist humans in teleoperation tasks with low visual information due to the medium or the lack of suitable sensors. We envision the proposed techniques can be applied to applications that include but are not limited to: blind grasping and object manipulation, underwater object localization and recognition, autonomous palpation in telesurgery, and navigation by visual-tactile fusion.

\section{Acknowledgement}
This material is based upon work supported by the National Science Foundation under Grant NSF NRI \#1925194 and \#2140612. Any opinions, findings, and conclusions or recommendations expressed in this material are those of the author(s) and do not necessarily reflect the views of the NSF.

{\revcolor{red}Chenxi Xiao and Juan Wachs contributed to the design, implementation and experiments of whisker sensor, exploration policy, and classification algorithms. Shujia Xu and Wenzhuo Wu contributed to whisker sensor fabrication, and motor testbed experiments.}

{\revcolor{red}
\normalsize
\section{Appendix} \label{sec:sp}
\subsection{Additional Results on Efficiency Characterization}

\begin{figure}[htb]
    \centering
    \subfloat[Scene \textcircled{2}, $U^S$]{\includegraphics[width=0.48\linewidth]{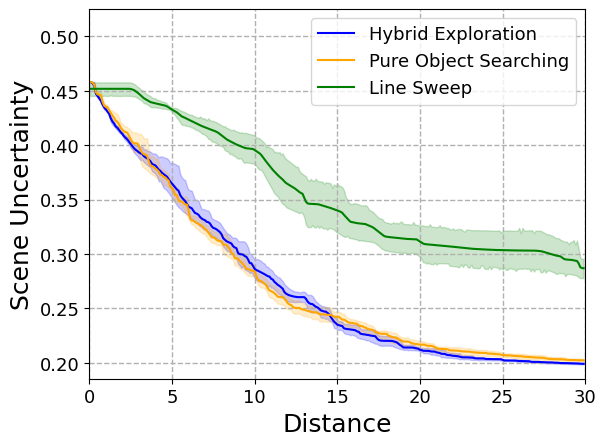}}
    \subfloat[Scene \textcircled{2}, $U^C$]{\includegraphics[width=0.48\linewidth]{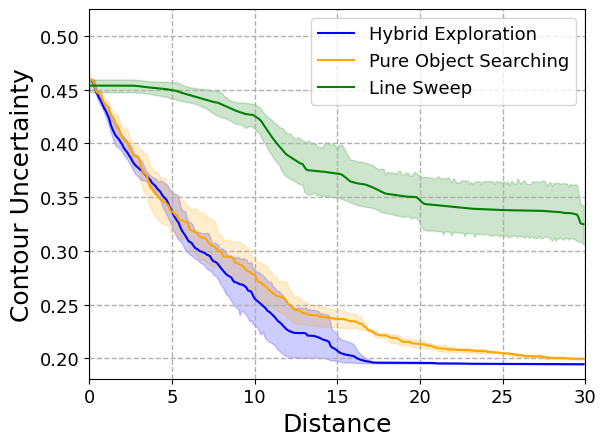}}
    
    \subfloat[Scene \textcircled{3}, $U^S$]{\includegraphics[width=0.48\linewidth]{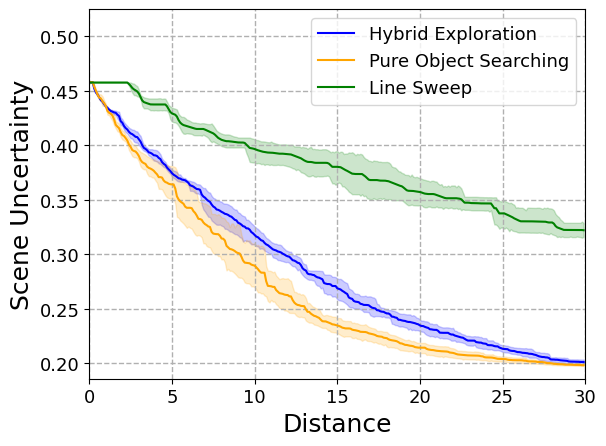}}
    \subfloat[Scene \textcircled{3}, $U^C$]{\includegraphics[width=0.48\linewidth]{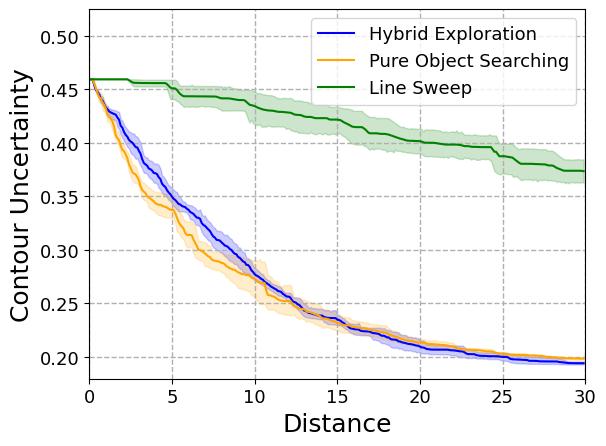}}
    
    \subfloat[Scene \textcircled{5}, $U^S$]{\includegraphics[width=0.48\linewidth]{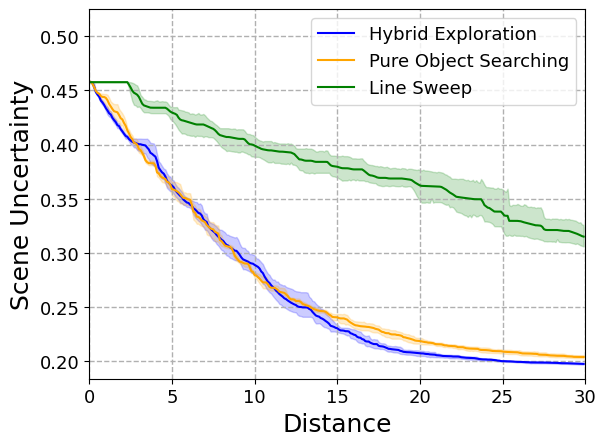}}
    \subfloat[Scene \textcircled{5}, $U^C$]{\includegraphics[width=0.48\linewidth]{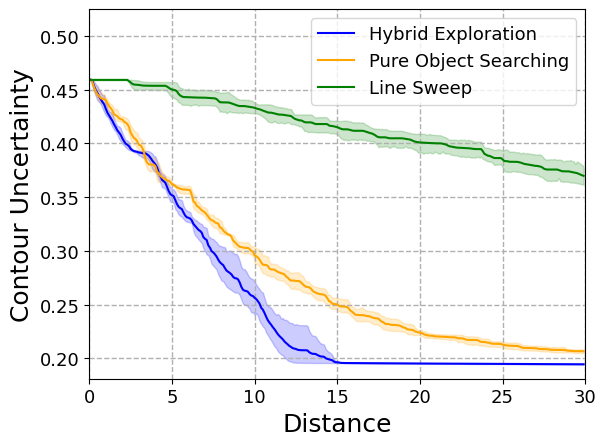}}
    
    \subfloat[Scene \textcircled{6}, $U^S$]{\includegraphics[width=0.48\linewidth]{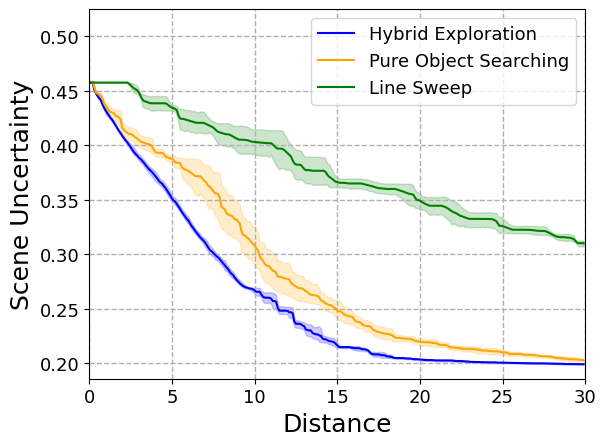}}
    \subfloat[Scene \textcircled{6}, $U^C$]{\includegraphics[width=0.48\linewidth]{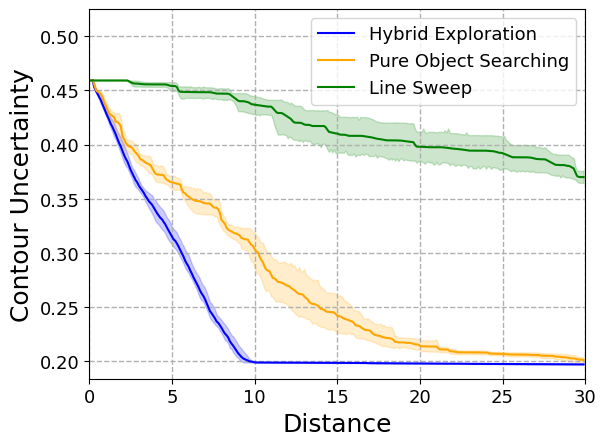}}
    \caption{Evaluation metric $U^S$ and $U^C$ versus the total travel distance in scenes \textcircled{2}\textcircled{3}\textcircled{5}\textcircled{6}. The shadowed regions are the 95\% confidence intervals.}
    \label{fig:metric_curve_supple}
\end{figure}

\captionsetup[subfigure]{labelformat=empty}
\begin{figure}[htb]
    \centering
    \subfloat[Scene \textcircled{2}]{\includegraphics[width=0.23\linewidth]{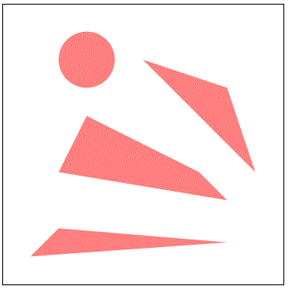}}
    \subfloat[(1)]{\includegraphics[width=0.23\linewidth,trim=40 40 40 40,clip]{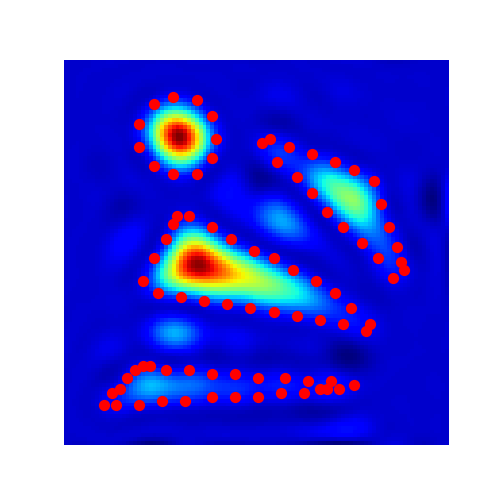}}
    \subfloat[(2)]{\includegraphics[width=0.23\linewidth,trim=40 40 40 40,clip]{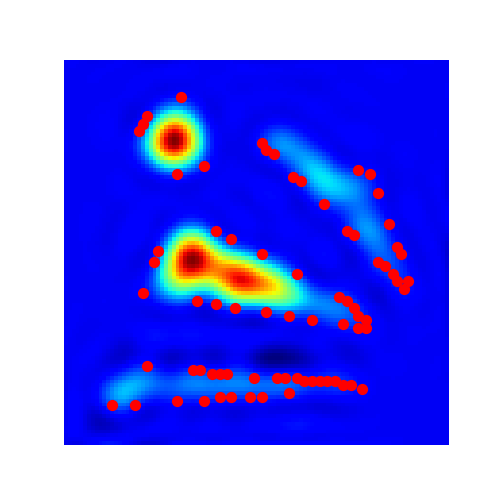}}
    \subfloat[(3)]{\includegraphics[width=0.23\linewidth,trim=40 40 40 40,clip]{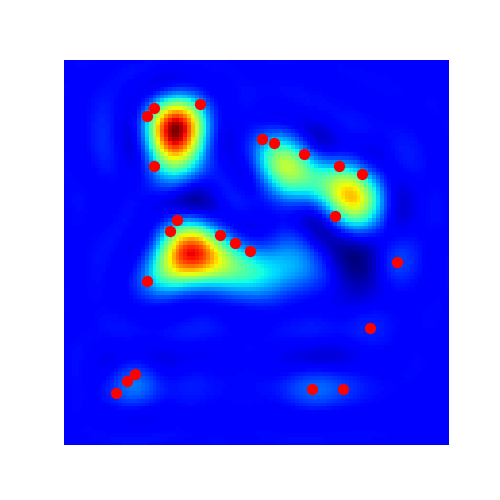}}

    \subfloat[Scene \textcircled{3}]{\includegraphics[width=0.23\linewidth]{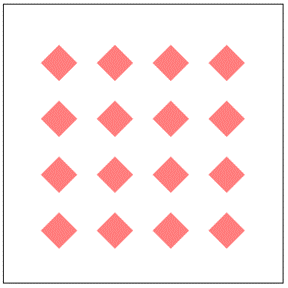}}
    \subfloat[(1)]{\includegraphics[width=0.23\linewidth,trim=40 40 40 40,clip]{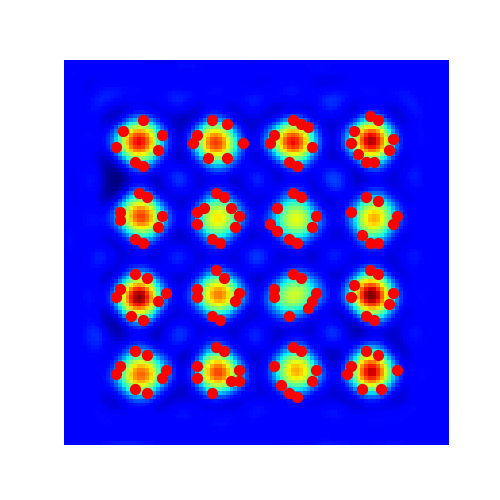}}
    \subfloat[(2)]{\includegraphics[width=0.23\linewidth,trim=40 40 40 40,clip]{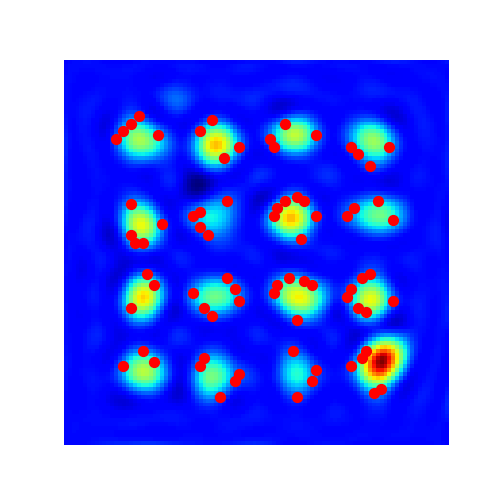}}
    \subfloat[(3)]{\includegraphics[width=0.23\linewidth,trim=40 40 40 40,clip]{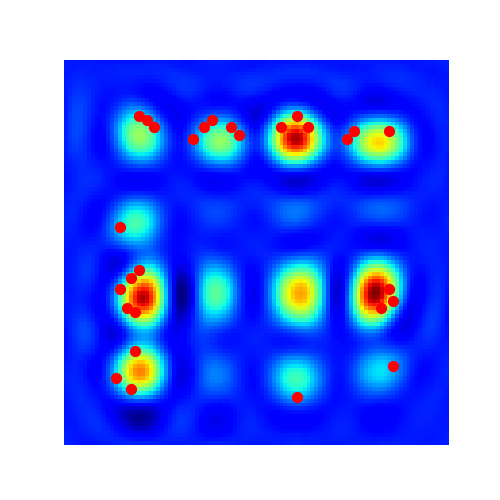}}

    \subfloat[Scene \textcircled{5}]{\includegraphics[width=0.23\linewidth]{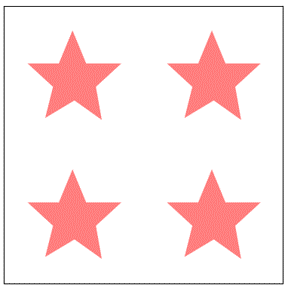}}
    \subfloat[(1)]{\includegraphics[width=0.23\linewidth,trim=40 40 40 40,clip]{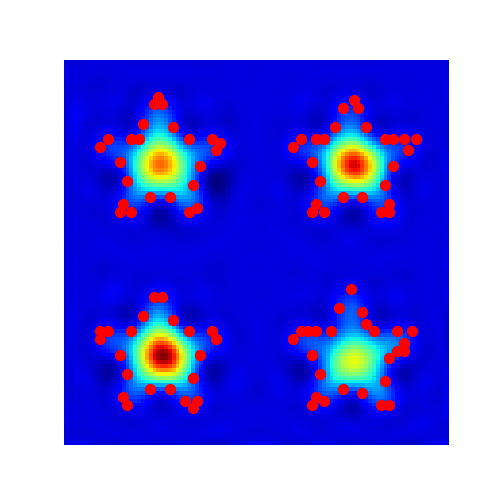}}
    \subfloat[(2)]{\includegraphics[width=0.23\linewidth,trim=40 40 40 40,clip]{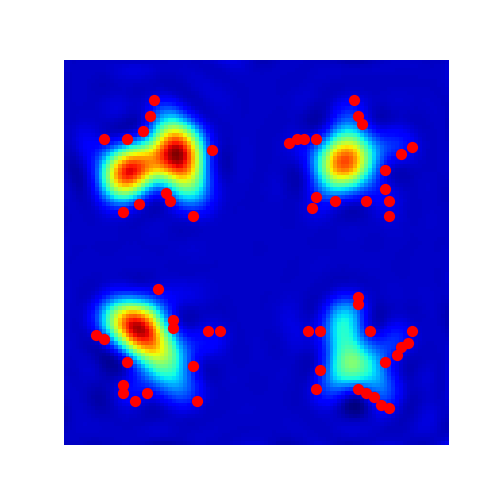}}
    \subfloat[(3)]{\includegraphics[width=0.23\linewidth,trim=40 40 40 40,clip]{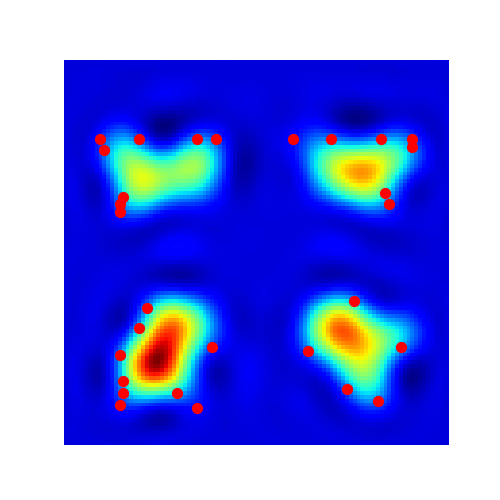}}
    
    \subfloat[Scene \textcircled{6}]{\includegraphics[width=0.23\linewidth]{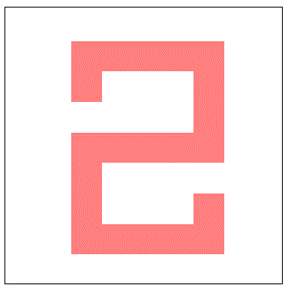}}
    \subfloat[(1)]{\includegraphics[width=0.23\linewidth,trim=40 40 40 40,clip]{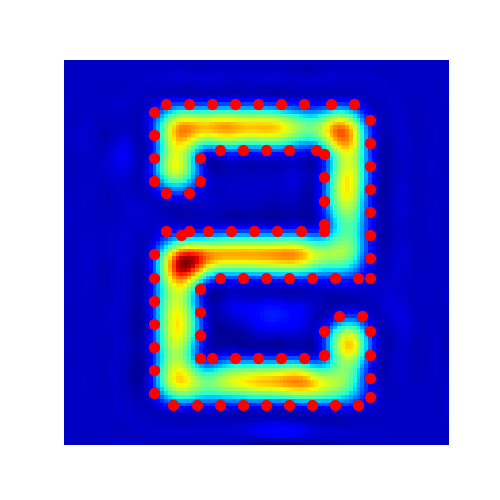}}
    \subfloat[(2)]{\includegraphics[width=0.23\linewidth,trim=40 40 40 40,clip]{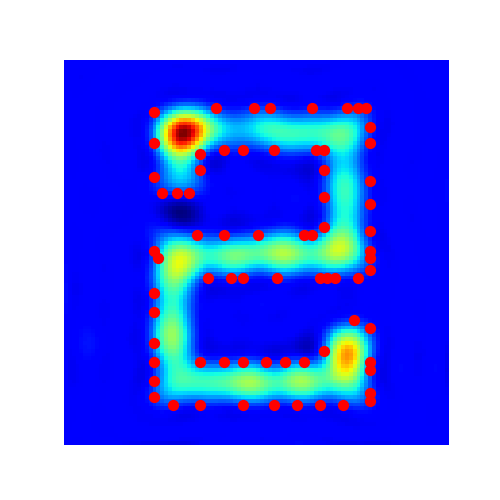}}
    \subfloat[(3)]{\includegraphics[width=0.23\linewidth,trim=40 40 40 40,clip]{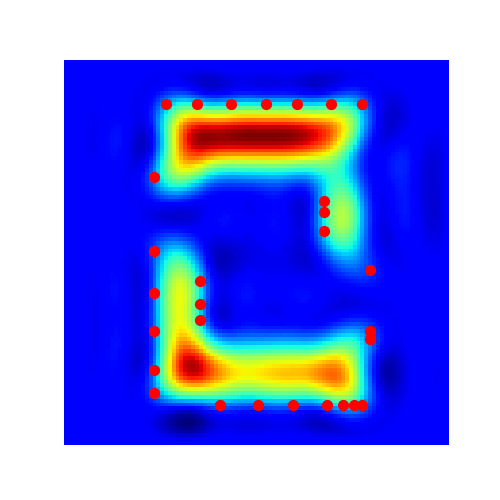}}

    \caption{Qualitative comparison of the exploration results in scenes \textcircled{2}\textcircled{4}\textcircled{5}\textcircled{6}. The evaluation protocol follows Fig.~\ref{fig:qual_30}.}
    \label{fig:qual_30_2}
\end{figure}
\captionsetup[subfigure]{labelformat=parens,aboveskip=0pt}

The benchmark results for additional scenes in Fig.~\ref{fig:reconstruct} are provided, which are obtained under the same protocol defined in Sec.~\ref{sec:exp_eff}. The results are demonstrated in Fig.~\ref{fig:metric_curve_supple} and Fig.~\ref{fig:qual_30_2}. In all scenes, both the hybrid policy and the object searching policy outperformed the line sweep baseline in both metrics. This is because additional traveling distance was induced by moving back and forth, and also because the baseline policy was not able to explore the regions occluded.  Also similar to the results of scene \textcircled{1} and \textcircled{4}, the hybrid policy outperformed pure object searching policy in the contour uncertainty metric $U^C$. The gap between the hybrid policy and pure object searching baseline is observed to be correlated to the scene layout.  When the object is small and scattered in multiple locations, the object searching policy plays a more significant role. This can be seen in the results of scene \textcircled{3}, in which the gap is relatively small. When the object dimension is relatively large, the advantage of the hybrid policy is significant due to the need of tracing object contour consistently, as demonstrated in the results of all other scenes.

The obtained contact points, and the predicted values of the occupancy function $f$ are demonstrated in Fig.~\ref{fig:qual_30_2}. The same conclusion can be reached as Fig.~\ref{fig:qual_30}. For our hybrid policy, the contact points were evenly distributed on the object surface, resulting in better prediction of the occupancy function than the other two approaches. These features were not found in other results, in which contact points were unevenly scattered, and did not fully convey the necessary information about the complete shape of objects.

\subsection{Long-term Characterization on Drifting}

\begin{figure}[htb]
    \centering
    \includegraphics[width=0.9\linewidth]{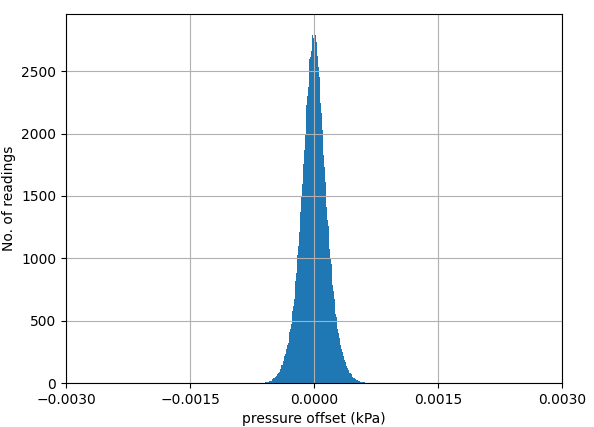}
    \caption{Bar plot of the output data distribution of channel 1. }
    \label{fig:longterm1}
\end{figure}

\begin{figure}[htb]
    \centering
    \includegraphics[width=0.9\linewidth]{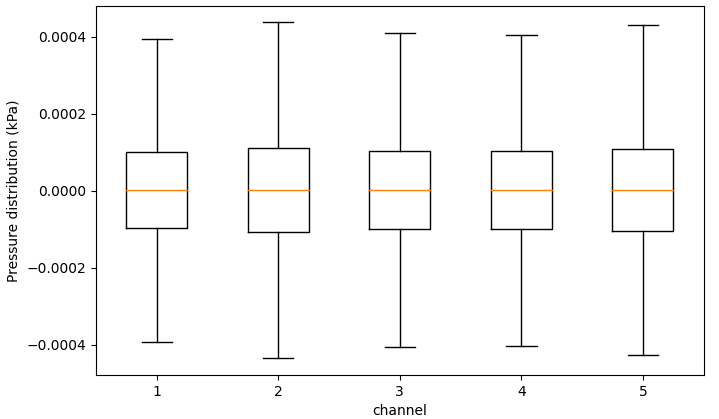}
    \caption{Box plot of the pressure distribution among all 5 channels.}
    \label{fig:longterm2}
\end{figure}

The sensor's output data were collected for approximately three and a half hours. This was done by continuously keeping the sensor still in a contact-free state. The goal was to characterize whether the drifting issue has been mitigated in the long term. For each whisker channel, it included 668,011 data points. The statistics of the data is visualized in Fig.~\ref{fig:longterm1} and Fig.~\ref{fig:longterm2}. For an individual channel (i.e., channel 1 in Fig.~\ref{fig:longterm2}), the sensor's output was approximated as a zero-mean normal distribution (actual mean: $9.5\times10^{-7}$ kPa), with the standard deviation $2.2\times10^{-4}$ kPa. This indicates that the drifting issue has been mitigated by the proposed approach.  

We also visualize the output distribution across channels by a box plot, as shown in Fig.~\ref{fig:longterm2}. Although minor differences exist in the variance of distributions, the drifting effect (i.e., the mean value, as the orange line) is observed to be around zero in all channels.

}

\small
\printbibliography

\newpage
\begin{minipage}[t][]{\columnwidth}
\vspace{-1.2cm}
\begin{IEEEbiography}
    [{\includegraphics[width=1.05in, height=1.3in]{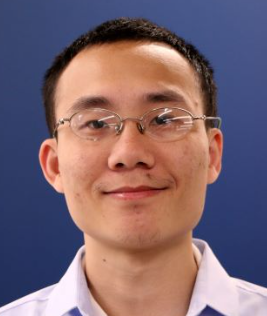}}]{Chenxi Xiao} received his BS and MS degrees in Electrical Engineering from the Northwestern Polytechnical University, Xi’an, China. He is currently working towards his Ph.D. degree in the School of Industrial Engineering at Purdue University.  He is also a research assistant at the Intelligent Systems and Assistive Technologies (ISAT) advised by Prof. Juan Wachs. His research interests include robot tactile exploration and manipulation, tactile sensors, and transfer learning.
\end{IEEEbiography}
\vspace{-1cm}
\begin{IEEEbiography}
    [{\includegraphics[width=1.05in, height=1.38in]{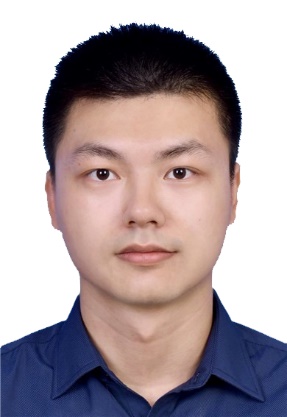}}]{Shujia Xu}
received his BS degree in Mechanical Engineering in 2015 form South China University of Technology (SCUT) and his MS degree in Biomedical Engineering in 2018 from Sun Yat-Sen University (SYSU). He is currently a PhD student in Industrial Engineering at Purdue University under the supervision of Prof. Wenzhuo Wu. His research interests include materials and manufacturing innovations of wearable electronics for human-integrated applications, including healthcare, human-machine interface, VR, and AI.
\end{IEEEbiography}
\vspace{-1cm}
\begin{IEEEbiography}
    [{\includegraphics[width=1.05in, height=1.3in]{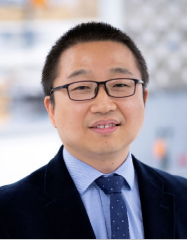}}]{Dr. Wenzhuo Wu}
 (Senior Member, IEEE) received his BS in Electrical Engineering in 2005 from the University of Science and Technology of China, Hefei, and his ME in Electrical and Computer Engineering from the National University of Singapore in 2008. Dr. Wu received his Ph.D. from Georgia Institute of Technology in Materials Science and Engineering in 2013. He is currently the Ravi and Eleanor Talwar Rising Star Associate Professor in the School of Industrial Engineering at Purdue University. His research interests include the design, manufacturing, and integration of nanomaterials for applications in energy, electronics, optoelectronics, and wearable devices. 
\end{IEEEbiography}
\vspace{-1cm}

\begin{IEEEbiography}
    [{\includegraphics[width=1.05in, height=1.3in]{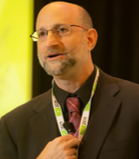}}]{Dr. Juan Wachs} (Senior Member, IEEE) is a University Scholar, Full Professor in the School of Industrial Engineering at Purdue University, Professor of Biomedical Engineering (by courtesy) and an Adjunct Associate Professor of Surgery at IU School of Medicine. He is the director of the Intelligent Systems and Assistive Technologies (ISAT) Lab at Purdue, and he is affiliated with the Regenstrief Center for Healthcare Engineering. Dr. Wachs received his B.Ed.Tech in Electrical Education in ORT Academic College, at the Hebrew University of Jerusalem campus. His M.Sc and Ph.D in Industrial Engineering and Management from the Ben-Gurion University of the Negev, Israel. 
\end{IEEEbiography}
\end{minipage}

\end{document}